\documentclass[10pt,journal,compsoc]{IEEEtran}
\usepackage{cite}
\usepackage{url}
\usepackage{amsmath,amsfonts}
\usepackage{algorithm}
\usepackage{algorithmic}
\usepackage{array}
\usepackage[caption=false,font=normalsize,labelfont=sf,textfont=sf]{subfig}
\usepackage{textcomp}
\usepackage{stfloats}
\usepackage{url}
\usepackage{hyperref}
\usepackage{verbatim}
\usepackage{amsmath}
\usepackage{amssymb}
\usepackage{booktabs}
\usepackage{multirow}
\usepackage{colortbl}
\usepackage[switch]{lineno}
\usepackage{graphicx}
\begin{document}

\title{Continual Adapter Tuning with Semantic Shift Compensation for Class-Incremental Learning}


\author{Qinhao Zhou, Yuwen Tan, Boqing Gong, Xiang Xiang$^*$\thanks{\indent Qinhao Zhou, Yuwen Tan, Xiang Xiang are all with the HUST AI \& Visual Learning Lab (HAIV Lab), Huazhong University of Science and Technology (HUST), Wuhan, Hubei 430074, China, while performing this work.\\ \indent Yuwen Tan and Boqing Gong are now both with the Department of Computer Science, Boston University, Boston, MA 02215, USA.\\ \indent Qinhao Zhou is now with ByteDance, Inc.\\ \indent Correspondence to Xiang Xiang (e-mail: \url{xex@hust.edu.cn}).}}

%
%

\markboth{Journal of \LaTeX\ Class Files}%
{Shell \MakeLowercase{\textit{et al.}}: Bare Demo of IEEEtran.cls for Computer Society Journals}
%



\IEEEtitleabstractindextext{%
\begin{abstract}
Class-incremental learning (CIL) aims to enable models to incrementally learn new classes while overcoming catastrophic forgetting of old classes. The introduction of pre-trained models has brought new tuning paradigms to CIL. In this paper, we first revisit different parameter-efficient tuning (PET) methods within the context of continual learning. It has been observed that continuous adapter tuning with local classifier training demonstrates superiority over prompt-based methods, even without parameter expansion and complex fusion operations during each learning session. Inspired by this, we propose incrementally tuning the shared adapter without imposing parameter update constraints, enhancing the learning capacity of the backbone.   We estimate the semantic shift of old prototypes without access to past image samples and update stored prototypes session by session. Additionally, we employ feature sampling from stored prototypes to retrain a unified classifier, further improving its performance. Our proposed method eliminates model expansion and avoids retaining any image samples. It surpasses previous pre-trained model-based CIL methods and demonstrates remarkable continual learning capabilities. Experimental results on several CIL benchmarks validate the effectiveness of our proposed method, achieving state-of-the-art (SOTA) performance. 
\end{abstract}

}

\maketitle

\IEEEdisplaynontitleabstractindextext

%
\IEEEpeerreviewmaketitle

\IEEEraisesectionheading{\section{Introduction}\label{sec:introduction}}



\IEEEPARstart{I}n traditional deep learning, the model can access all the data at once and learning is performed on a static dataset. However, in real-life applications, data usually arrives in a streaming format with new classes, requiring the model to learn continuously, known as class-incremental learning (CIL). The primary objective of CIL is to enable the model to learn continuously from non-stationary data streams, facilitating adaptation to new classes and mitigating catastrophic forgetting \cite{french1999catastrophic} of past learned classes. Traditional CIL methods \cite{rebuffi2017icarl,wang2022foster,zhu2021prototype} have been devoted to alleviating catastrophic forgetting. Those methods can be mainly divided into replay-based \cite{bang2021rainbow, chaudhry2018riemannian, rebuffi2017icarl}, regularization-based \cite{kirkpatrick2017overcoming, zenke2017continual, aljundi2018memory}, and isolation-based methods \cite{serra2018overcoming, mallya2018packnet, mallya2018piggyback}. However, all these methods assume that models are trained from scratch while ignoring the generalization ability of a strong pre-trained model \cite {dosovitskiy2020image,radford2021learning} in the CIL. 


Recently, pre-trained vision transformer models \cite{dosovitskiy2020image} have demonstrated excellent performance on various vision tasks. It has also been explored in the field of CIL and continues to receive considerable attention \cite{zhou2023revisiting, wang2022learning, wang2022dualprompt,zhang2023slca,zhou2024expandable,huang2024ovor,zhou2024continual}. Due to the powerful representation capabilities of pre-trained models, CIL methods based on pre-trained models achieve significant performance improvements compared to traditional
 SOTA methods which are trained from scratch. CIL with a pre-trained model typically fixes the pre-trained model to retain the generalizability and adds a few additional training parameters such as adapter \cite{chen2022adaptformer}, prompt \cite{jia2022visual} and SSF \cite{lian2022scaling}, which is referred to as parameter-efficient tuning (PET). 
 Simple but effective PET methods \cite{hu2021lora,jia2022visual,chen2022adaptformer,zhang2023llama} can significantly boost the performance of pre-trained models in downstream tasks but also in continual learning scenarios.  

\begin{figure}[t]
\centering
\includegraphics[scale=0.31]{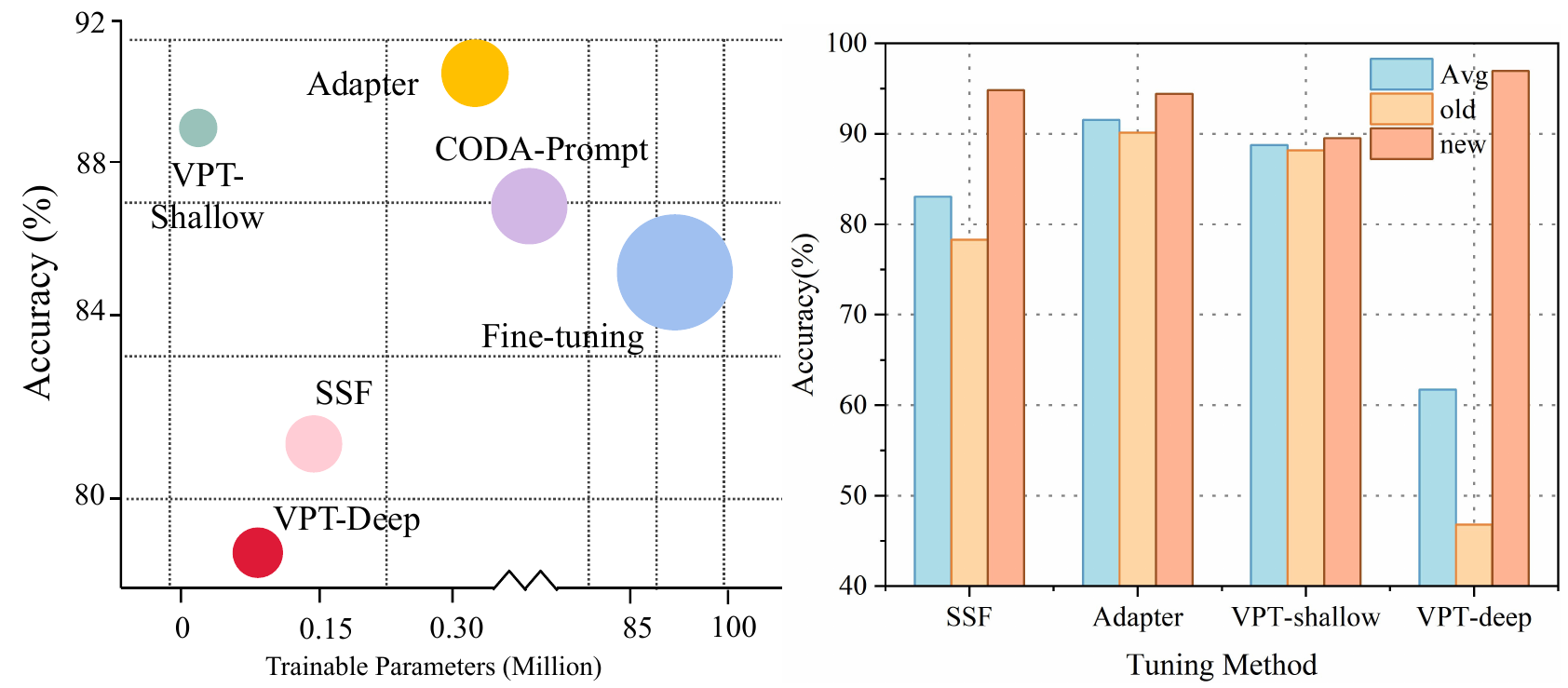}        
\vspace{-2mm}
\caption{Comparison of different parameter-efficient tuning CIL baselines on CIFAR100 dataset. \textbf{Left:} The relationship between the average accuracy of the incremental sessions and the number of tunable parameters. \textbf{Right:} The average performance of old classes and new classes for each PET method. This figure show that Adapter tuning performs best and is more stable across datasets compared to other PET methods.}
\label{fig1}
\centering
\end{figure}

Inspired by language-based intelligence, current research in  CIL is primarily focused on the prompt-based method such as L2P\cite{zhou2022learning}, DualPrompt\cite{wang2022dualprompt}, and CODAprompt \cite{smith2023coda}. Typically, these approaches require the construction of a pool of task-specific prompts during the training phase which increases storage overhead. Additionally, selecting prompts during the testing stage incurs additional computational costs. Other PET methods as well as fully fine-tuning are still in exploration in the context of CIL. Recently, SLCA \cite{zhang2023slca}
proposes fine-tuning the entire ViT and classifier incrementally with different learning rates. However, fine-tuning the entire pre-trained model requires substantial computational resources. In addition, APER \cite{zhou2023revisiting} initially explores the application of other PET methods in CIL using first-session adaptation and branch fusion.  Training in the first stage and subsequently freezing the model \cite{zhou2023revisiting,panos2023first} can reduce training time but result in lower accuracy for subsequent new classes. Our linear probing results reveal that the first-session adaptation is insufficient when there is a significant domain discrepancy between downstream data and the pre-trained model. 
EASE \cite{zhou2024expandable} propose an expandable adapter structure that yields substantial performance improvements over ADAM \cite{zhou2023revisiting}. However, it requires continuous expansion of the model architecture and
extra storage to save the adapters.

In this paper, we first revisit different PET methods within the CIL paradigm. As shown in Fig.~\ref{fig1}, we observe that adapter tuning \cite{chen2022adaptformer} is a better continual learner than prompt-tuning \cite{jia2022visual} and SSF-tuning \cite{lian2022scaling}. When progressively fine-tuning the prompt and SSF parameters, the forgetting of old classes is catastrophic. In comparison, adapter tuning effectively balances learning new classes and maintaining performance in old classes. Unlike prompt-based methods, which require constructing a prompt pool, adapter tuning avoids catastrophic forgetting even sharing the same parameters across learning sessions. Additionally, the adapter balances the number of tuning parameters and model performance compared to fully fine-tuning. Moreover, unlike previous methods that use feature distillation loss to restrict changes in shared parameters as part of overall loss, we analyze that tuning with constraints hinders continual learning from the perspective of parameter sensitivity. Therefore, we train the adapter and task-specific classifier without parameter regularization in each session, allowing for greater plasticity in learning new classes. 
In addition, we find that continuously fine-tuning the adapter is also effective in other CIL settings, such as few-shot class-incremental learning (FSCIL) \cite{tao2020few}  and hierarchical class-incremental learning (HCIL) \cite{zhou2023hierarchical}.

Furthermore, as we only train the local classifier in each learning session, we propose to adopt a new classifier retraining method  \cite{zhu2021prototype,zhang2023slca,tang2023prompt} to further improve the CIL performance. First, we implicitly compute the semantic shift of previous prototypes by leveraging the shift of current task prototypes. Different from SDC\cite{yu2020semantic} which estimates the change of old prototypes by calculating the shift of current samples between different learning sessions, we estimate the shift by calculating the change of prototypes between different tasks, thereby reducing computational costs.  Then, we sample multiple features according to the updated prototypes to retrain the classifier which is more effective than previous methods without semantic shift estimation. The advantages of our proposed method can be summarized as follows: 1) Fine-tuning adapters significantly reduces training costs and improves learning efficiency compared to training the whole model; 2) We do not need to retain any image samples; 3) The accuracy for new classes is relatively high which verifies the continual learning capacity of the model. Extensive experimental results on seven CIL benchmarks demonstrate the superiority of the proposed simple but effective methods which achieves the SOTA; We also conduct experiments on FSCIL and HCIL setting and gain excellent performance as well.

 
In summary, our proposed learning framework has the following main contributions: (1) Different from various devotion into the prompt-based methods for CIL, we discover that incrementally tuning adapter is a better continual learner even without constructing an adapter-pool compared with other PET methods;
(2) After each session adaptation with local classifier, we propose to retrain a unified classifier with the semantic shift compensated prototypes which can further improve the CIL performance; (3) We apply our method to FSCIL and HCIL settings, demonstrating superior performance compared to other approaches in these contexts. 
Compared to our CVPR 2024 version \cite{tan2024semantically}, in the contribution (2),  {\bf we propose a more efficient semantic shift estimation method}. Furthermore, in our experiments, we extend the experiments on fine-grained datasets to further verify the effectiveness of our method. Moreover, we conduct experiments on other incremental settings.

 The rest of this article is organized as follows. We first summarize the existing related work on CIL  in Section \ref{related work} and provide the preliminaries of the setting and method in Section \ref{Preliminary}. Then, we give a detailed description of the proposed method in Section \ref{Method}. To verify the effectiveness of our proposed method, experiments and ablation studies are given in Section \ref{Exp}. Finally, we conclude and discuss the future work in Section \ref{conclusion}.

\section{Related Work}
\label{related work}

\subsection{Traditional Class-incremental Learning}
Class-incremental learning requires the model to be continuously updated with new class instances while effectively retaining old knowledge \cite{zhou2023deep}. Traditional CIL methods typically train the model from scratch, primarily focusing on mitigating the forgetting of old classes. To alleviate the forgetting, various data-centric approaches have been proposed, including storing past samples \cite{bang2021rainbow, chaudhry2018riemannian, rebuffi2017icarl, sun2023exemplar, zhu2021prototype, petit2023fetril} or synthesizing image samples with generative models \cite{shin2017continual, he2018exemplar, hu2019overcoming}. In addition to data-centric methods,  model-centric methods have also been explored. Model-centric methods include expanding the model parameters \cite{yan2021dynamically, wang2022foster, zhou2022model}, restricting parameter updates \cite{chaudhry2018riemannian, kirkpatrick2017overcoming, zenke2017continual, xiang2022coarse}, and parameter isolation \cite{serra2018overcoming, mallya2018packnet, mallya2018piggyback}. 
Moreover, beyond model-centric and data-centric perspectives, a growing number of methods emphasize the continuous fine-tuning of models with minimal forgetting, proposing various approaches at the algorithm level. For example,  knowledge distillation \cite{wen2024class, zuo2024hierarchical, gao2022r, bonato2024mind}, balancing batch normalization \cite{cha2023rebalancing}, feature space shift \cite{li2024fcs, he2024dyson}. These general model-agnostic CIL methods can also be effectively applied to the continuous fine-tuning process of pre-trained models to alleviate forgetting.


\subsection{Parameter-Efficient Tuning}
Parameter-Efficient Tuning can be considered as a transfer learning method. It refers to not performing full fine-tuning on a pre-trained model, instead inserting and fine-tuning specific sub-modules within the network. This fine-tuning paradigm does not modify the existing pre-trained model parameters and have effective transfer learning results in NLP \cite{houlsby2019parameter, lester2021power, li2021prefix, hu2021lora}. Recently, similar approaches have been applied to vision transformer models as well. AdaptFormer \cite{chen2022adaptformer} inserts lightweight modules after the MLP layers in the attention module and has been found to outperform full fine-tuning on action recognition benchmarks. Another PET approach SSF \cite{lian2022scaling} 
 surprisingly outperforms other methods in certain tasks even with a smaller number of parameters. Inspired by the prompt approach used in the language model, VPT \cite{jia2022visual} applies it to visual models and achieves impressive results across various downstream tasks while only introducing a small number of additional parameters. Furthermore, the prompt-based method has also been used in vision-language models \cite{radford2021learning, zhou2022learning, zhou2022conditional, zhang2022pointclip} to improve performance on various downstream tasks.

\subsection{Continual Learning on a Pre-trained Model}
The aforementioned CIL methods all involve training the model from scratch, while CIL with pre-trained model \cite{wu2022class,zhou2022learning,zhou2023revisiting,wang2022s, zhao2024continual} has gained much attention due to its strong feature representation ability. L2P \cite{zhou2022learning} utilizes the pre-trained model and learns a set of extra prompts dynamically to guide the model to solve corresponding tasks.  DualPrompt \cite{wang2022dualprompt} proposes to learn of two mutually unrelated prompt spaces: the general prompt and the expert prompt. It encodes task-invariant instructions and task-specific instructions, respectively. CODAPrompt \cite{smith2023coda} introduces a decomposed attention-based continual learning prompting method, which offers a larger learning capacity than existing prompt-based methods \cite{zhou2022learning,wang2022dualprompt}. SLCA \cite{zhang2023slca} explores the fine-tuning paradigm of the pre-trained models, setting different learning rates for backbone and classifiers, and gains excellent performance. ADAM \cite{zhou2023revisiting} proposes to construct the classifier by merging the embedding of a pre-trained model and an adapted downstream model. LAE \cite{gao2023unified} proposes a unified framework that calibrates the adaptation speed of tuning modules and ensembles PET modules to accomplish predictions.  EASE
\cite{zhou2024expandable} proposes an expandable adapter ensemble approach, integrating it with a pre-trained model to achieve superior CIL performance. However, this method requires additional model size to save the trained adapters. Several methods \cite{wang2022s,zhou2023learning,yu2024boosting} have been proposed to enhance the incremental performance of the model by leveraging multi-modal information of the pre-trained vision-language model.

\subsection{Few-shot Class-incremental Learning and Hierarchical Class-incremental Learning }
In addition to the regular CIL setting, two particular CIL settings have garnered significant attention in recent years: few-shot class-incremental learning (FSCIL) and hierarchical class-incremental learning (HCIL). TOPIC \cite{tao2020few} first proposes FSCIL and CEC \cite{zhang2021few} which decouples the encoder and classifier is a widely used baseline. FACT \cite{zhou2022forward} and ALICE \cite{peng2022few} employ pseudo-class data augmentation and consider the compatibility for future classes, which achieve significant performance improvements. Limit \cite{zhou2022few} and SPPR \cite{zhu2021self} leverage the base-stage data to construct few-shot settings, and then optimize the model to better adapt to the subsequent few-shot learning sessions. SAVC \cite{song2023learning} summarizes and builds upon previous work and leverages supervised contrastive learning and pseudo-class construction techniques, which achieve remarkable results. HIL extends the concept of CIL by taking into account the hierarchical relationships between classes. IIRC \cite{abdelsalam2021iirc} evaluates the performance of CIL methods in the context of HIL setting. Furthermore, other HIL settings have also been proposed in recent years \cite{liang2022balancing,zhou2023hierarchical,xiang2022coarse}. Since both FSCIL and HIL are extensions of CIL, we evaluate our method on these two extra settings to validate the generalizability. 

\begin{figure*}[t]
\centering
\includegraphics[scale=0.53]{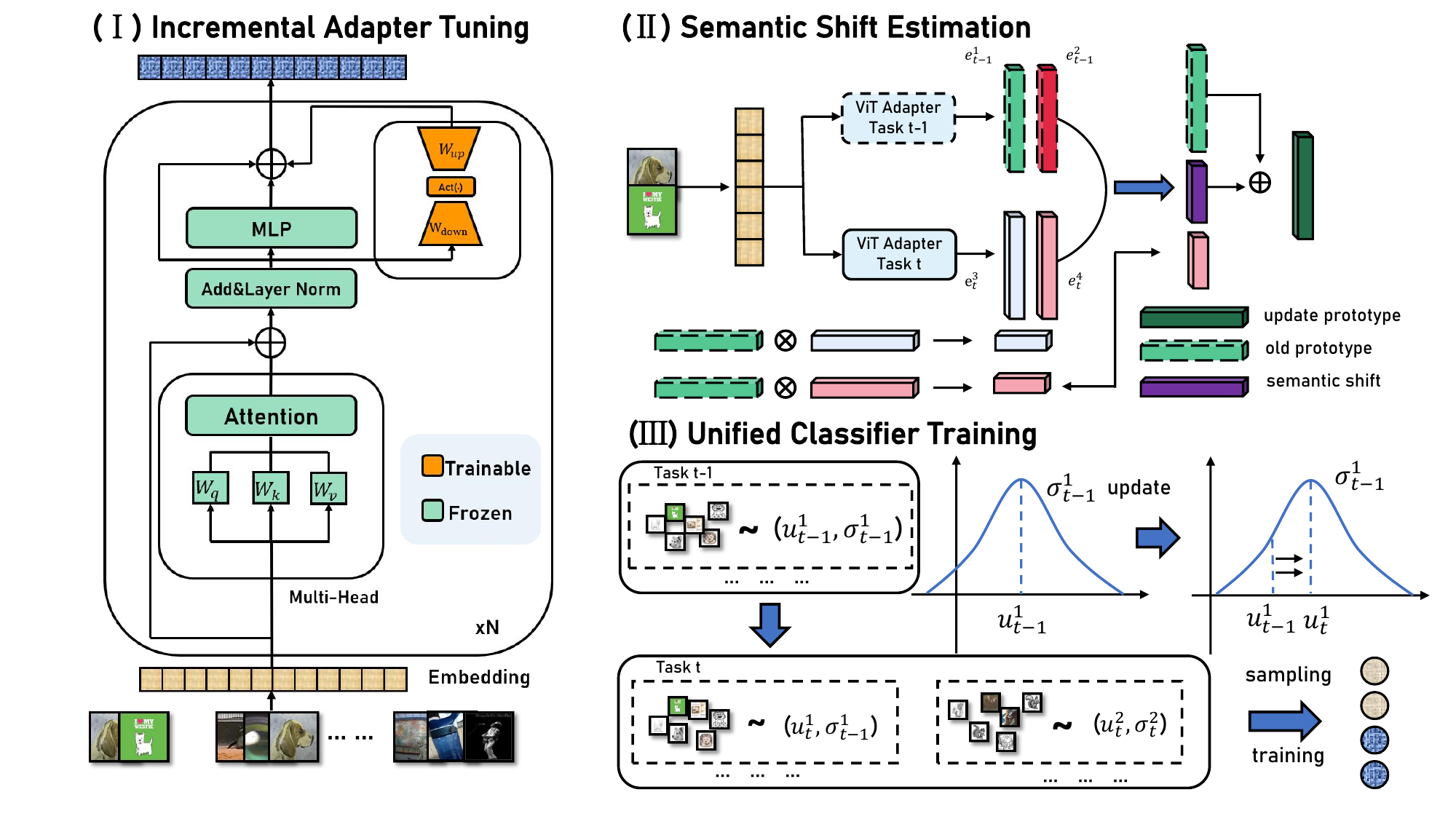} 
\vspace{-3mm}
\caption{The framework of our proposed method. \textbf{Left:} The illustration of the structure of ViT and adapter. The adapter and local classifier are incrementally trained in each session using the Eq.~\ref{loss}.  \textbf{Right:} The process of retraining the classifier with semantic shift estimation.}
\label{overall framework}
\vspace{-3mm}
\end{figure*}

\section{Preliminary}
\label{Preliminary}
 \textbf{Class-incremental learning formulation:} We first introduce the definition of CIL. Consider a neural network $\mathcal{M}_{\theta}=f_{\theta_{cls}}(\mathcal{F}_{\theta_{bne}}(\cdot))$ with trainable parameters $\theta=\{\theta_{bne}, \theta_{cls}\}$. $\mathcal{F}_{\theta_{bne}}$ represents the feature
extraction backbone which extracts features from input images and $ f_{\theta_{cls}} $ stands for the classification layer that projects feature representations to class predictions. In CIL setting, $\mathcal{M}_{\theta}$ needs to learn a series of sessions from training data $D_t=\{(x^t_1,y^t_1), (x^t_2,y^t_2),...\}, t=1,...,T$ and satisfy the condition $Y(i) \cap Y(j) = \emptyset ,i\neq j$, where $Y(i)$ represent the label set in session $i$. The goal of  $\mathcal{M}_{\theta}$ is to perform well on test sets that contain all the classes learned denoted as $\mathcal{Y}=Y(1) \cup ...Y(t) $  after $t$-$th$ session.

\textbf{Parameter-efficient tuning with Adapter:} An adapter is a bottleneck structure \cite{chen2022adaptformer} that can be incorporated into a pre-trained transformer-based network to facilitate transfer learning and enhance the performance of downstream tasks. An adapter typically consists of a downsampled MLP  layer $W_{down} \in \mathbb{R}^ {d \times \hat{d}}$, a non-linear activation function $\sigma$, and an upsampled MLP layer $W_{up} \in \mathbb{R}^{\hat{d} \times d}$. Denote the input as $x_i$, we formalize the adapter as  
\begin{equation}
out = x_i+ s\cdot\sigma(x_i* W_{down})* W_{up},
\end{equation}
where $*$ stands for the matrix multiplication, $\sigma$ denotes the activation function RELU, and $s$ denotes the scale factor.

\textbf{Parameter-efficient tuning with SSF:}  SSF \cite{lian2022scaling} modulates pre-trained models using scale and shift factors to align the feature distribution of downstream tasks. SSF inserts its layers in each transformer operation. Suppose $x_i$ is the output  of one of the modules, SSF can be  represented as 
\begin{equation}
    y = \gamma \odot x_i + \beta,
\end{equation}
where $\gamma \in \mathbb{R}^d$ and $\beta \in \mathbb{R}^d$ denote the scale and shift factor, respectively.  $\odot$ stands for Hadamard product.

\textbf{Parameter-efficient tuning with VPT:} Visual Prompt Tuning (VPT) inserts a small number of trainable parameters in the input space after the embedding layer \cite{jia2022visual}. It is called prompts and only these parameters will be updated in the fine-tuning process. Depending on the number of layers inserted, VPT can be categorized as VPT-shallow and VPT-deep. Suppose $P=\{p^k \in R^d| 1 \leq k \leq n\}$ and the input  embedding is $x$, VPT will combine $x$ with $P$ as 
\begin{equation}
    x' = [x, P],
\end{equation}
where $n$ is the number of prompts and the $x'$  will be passed into subsequent blocks.

\section{Methodology}
\label{Method}

\subsection{Adapter-tuning without parameter constraints}

 Most of the work based on pre-trained models focuses on how to apply the prompt-tuning strategies to the CIL paradigm. However, tuning the same prompt parameters across each learning session will cause catastrophic forgetting.  As shown in Fig.~\ref{fig1}, when progressively training the shared extra module while keeping the pre-trained model fixed, the adapter demonstrates its superiority over other tuning methods such as prompt-tuning and SSF. \textit{Fine-tuning the shared adapter incrementally seems to well balance the learning of new classes and old-knowledge retaining.} Based on this observation, we delve deeper into incremental adapter tuning and use it as our baseline.  The whole framework of the proposed method is shown in Fig.~\ref{overall framework}. Some methods \cite{panos2023first,zhao2023revisit} adopt the first-session adaption and then fix the backbone. In addition,  previous methods often utilize knowledge distillation \cite{hinton2015distilling} (KD) loss to restrict parameter changes of the feature extractor to mitigate forgetting. Totally different from earlier methods \cite{li2017learning,rebuffi2017icarl, kirkpatrick2017overcoming}, we propose that the shared adapter should be tuned incrementally without parameter constraints. Next, we will provide a detailed description of the proposed baseline and offer a reasonable explanation and analysis.

\textbf{Implementation of adapter-based baselines:} During incremental training sessions, only adapter and classifier layers are updated, and the pre-trained ViT model is frozen. As the cosine classifier has shown great success in CIL, we follow ALICE \cite{peng2022few} to use the cosine classifier with a margin. The margin hyper-parameter could also be used as a balance factor to decide between learning new classes and retaining old knowledge. The training loss can be formulated as follows:
\begin{equation}
   \mathcal{L}^t=-\frac{1}{N^t}\sum_{j=1}^{N^t} log \frac{e^{s(cos \theta_j^i-m)}}{e^{s(cos \theta_j^i-m)}+\sum_{c=1 }^{Y(t)-\{i\}}e^{s(cos \theta_j^c)}}
\label{loss}
\end{equation}
where $cos \theta_j^i= \frac{w_i*f_j}{|| w_i ||* || f_j || } $, $N^t$ denotes the number of training samples of the current session, $s$ and $m$ represent the scale factor and margin factor, respectively.

As we do not retain any image samples, the gradients computed during the optimization of current samples not only affect the newly trained classifiers but also have an impact on the previously learned classifiers. The forgetting of the classifier is significant when no samples are retained. Thus, we follow previous work \cite{zhang2023slca,wang2023isolation,gao2023unified} to adopt the local training loss where we only compute the loss between current logits and labels and hinder the gradient updates of the previous classifier which alleviates the classifier forgetting.

\begin{figure}[t]
\centering
\includegraphics[scale=0.22]{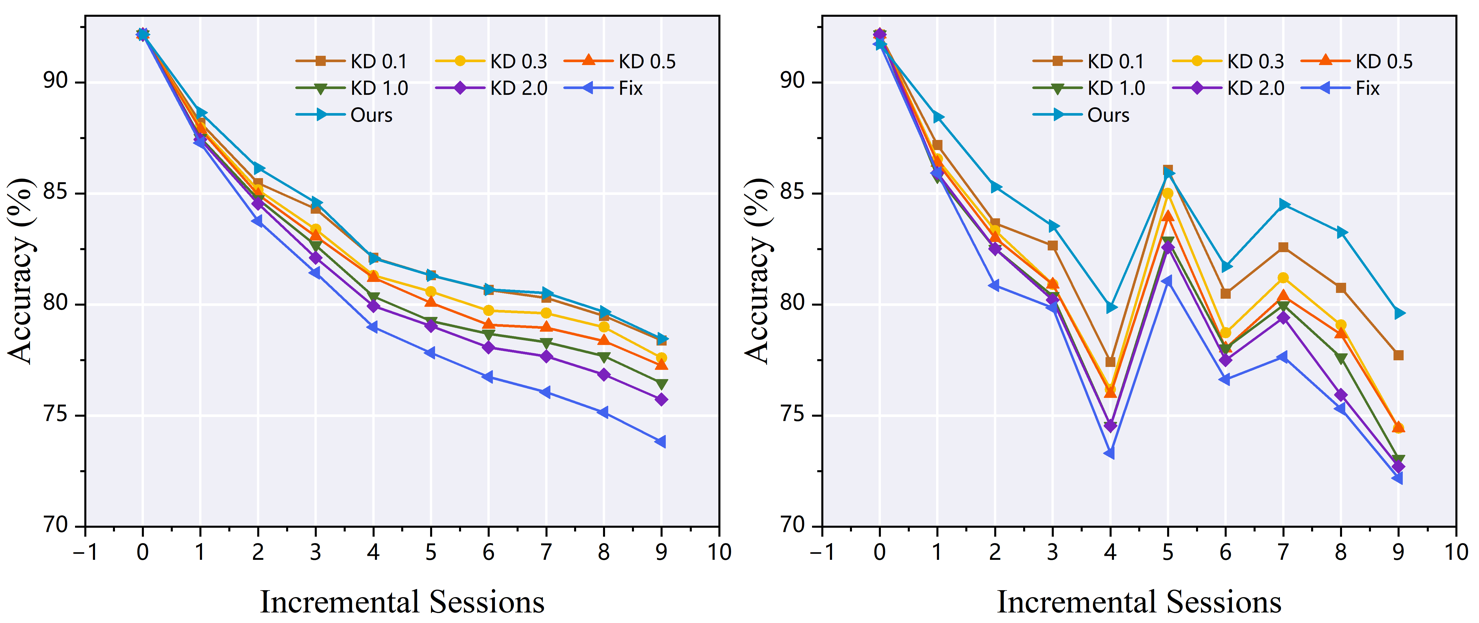}        
\vspace{-2mm}
\caption{Comparison of the performance on ImageNetR dataset with different extent of parameter constraints. \textbf{Left:} The overall accuracy of each session. \textbf{Right:} The accuracy of new classes.}
\label{fig4}
\centering
\end{figure}

\begin{figure}[t]
\centering
\includegraphics[scale=0.22]{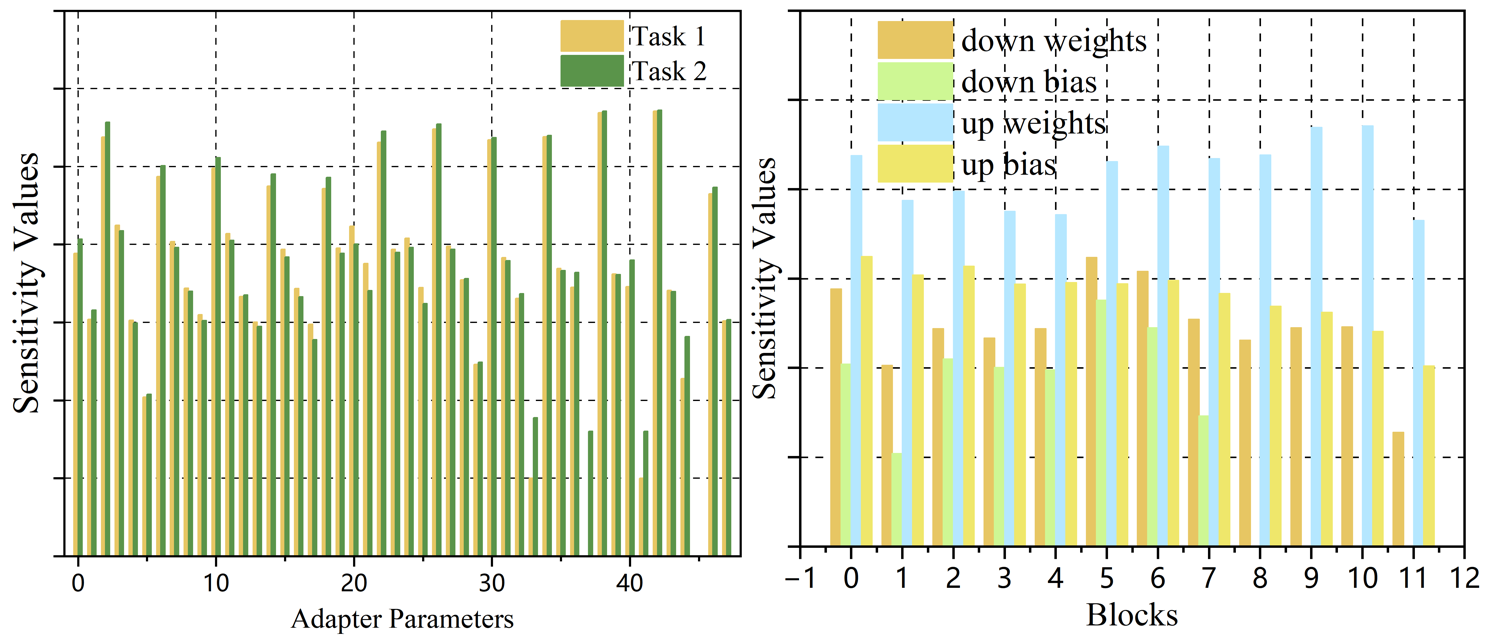}        
\vspace{-2mm}
\caption{ Parameter sensitivity analysis on the ImageNetR dataset. \textbf{Left:} The parameter sensitiveness of two incremental tasks.  \textbf{Right:} The sensitiveness of different parameters in one task.}
\label{fig3}
\vspace{-2mm}
\centering
\end{figure}

\textbf{Analysis of the adapter-based baseline:} We will analyze why the adapter shows its superiority in the CIL over other PET methods, and why we choose to incrementally tune the shared adapter without parameter constraints.


First, we elaborate on why incrementally tuning the adapter is better in the context of CIL. By utilizing the residual structure, the adapter can retain the generalization capabilities from the pre-trained model while adapting to new tasks. The incremental tuning of the adapter exhibits a cumulative learning capability, where the representational capacity of the adapter is further enhanced as the learning sessions progress. In contrast, both SSF and prompt tuning have limitations when it comes to handling CIL. These methods suffer from overfitting to the current distribution. When the shared parameters excessively overfit each current task, the model gradually loses its generalization ability which is harmful for training a unified model for CIL.
 Then, we try to utilize KD loss to implicitly limit parameter updates and adjust the weighting factor. As shown in Fig.~\ref{fig4}, the results demonstrate that unconstrained training is more beneficial for new classes learning and improving overall performance. Based on this observation, we propose our proposition from the perspective of parameter sensitivity.\\

\noindent \textbf{Proposition 1}: \textit{Confining the change of parameters of previous tasks hinders the plasticity of new classes due to the similarity of parameter sensitivity among tasks.}

Proof: Given the parameter set $\theta=\{\theta_1,\theta_2,...\theta_N \}$ and training set $D_t=(X_t,Y_t)$ in $t$-$th$ session, the definition of parameter sensitivity \cite{he2023sensitivity,zhao2023revisit} is defined as
\begin{equation}
   s_i^t=\mathcal{L}(X_t,Y_t|\theta_i)- \mathcal{L}(X_t,Y_t|\theta_i^*),
\end{equation}
where $\theta_i^*=\theta_i+\Delta \theta_i$ and $\mathcal{L}$ denotes the optimized loss in the classification task. We use the first-order Taylor expansion, and the parameter sensitivity can be rewritten as follows: 
\begin{equation}
   s_i=-g_i\Delta \theta_i= -\frac{\delta \mathcal{L}}{\delta \theta_i}* \Delta \theta_i,
\end{equation}
as $\Delta \theta_i$ denotes the update after the training process, we follow the work \cite{he2023sensitivity}  to use the one-step update to approximate the  $\Delta \theta_i= \epsilon \frac{\delta \mathcal{L}}{\delta \theta_i} $. Therefore, the parameter  can be approximately computed as  $s_i \approx -\epsilon (\frac{\delta \mathcal{L}}{\delta \theta_i})^2 $.  As shown in Fig.~\ref{fig3}, the sensitivity values of tuning parameters for two different sessions are nearly equal and the most sensitive parameters are always \textit{up weights}. This means that constraining the parameter update would hinder the learning of new classes and further impede the ability of the model for continual learning.  Furthermore, in the experimental section, we demonstrate the representative capacity of the adapter continued to strengthen through incremental tuning.


\begin{algorithm}
\caption{Semantic Shift Estimation of Prototypes}
\label{alg:semantic_shift}  
\hspace*{0.02in}{\bf Input:} Current session training data $D_t$, Old Class Prototypes $P_o$, Old Model $M_o$, New Model $M_n$.

\hspace*{0.02in}\vspace*{0.03in}{\bf Output:} 
Updated old class prototypes $P_o^n$ with semantic shift estimation.
\begin{algorithmic}[1]
\label{alg:ss}
\STATE Calculate the prototypes of new classes $P_n^o$ and $P_n^n$ on both the old model $M_o$ and the new model $M_n$ using the dataset $D_t$.

\STATE Calculate the gap $\varphi_i^{t-1\to t}$ of the new class prototype on the old model and the new model according to Eq.~\ref{eq 9}.
\FOR{$i=1$ to $O_n$}
    \FOR{$j=1$ to $N_n$}
    \STATE Calculate the cosine similarity $\alpha_{ij}$ of old class prototype $P_i$ and new class prototype $P_j$.
    \ENDFOR
\ENDFOR

\FOR{$i=1$ to $O_n$}
    \STATE Calculate the semantic shift $\widetilde{\Delta}_i^{t-1\to t}$ of old class $i$ according to Eq.~\ref{compte-mean}.
    \STATE Add the semantic shift $\widetilde{\Delta}_i^{t-1\to t}$ to the old class prototype and get $P_o^n$.
\ENDFOR
\end{algorithmic}
\end{algorithm}

\subsection{Semantic shift estimation without past samples}
Due to the selective updating of classifiers corresponding to the current task during training, the classifiers across different learning sessions are not fully aligned in the same feature space. To further optimize classifiers, we store the prototypes after training the backbone and local classifier. However, as the backbone is trained incrementally with new classes, the feature distribution of old classes changes. Retraining the classifier with the previous prototypes is sub-optimal. Since the feature representability of the backbone updates over time, using outdated features may not effectively retrain a unified classifier. To solve this problem, we update the feature distribution of old classes by computing the semantic shift over the learning process. Inspired by the previous work SDC \cite{yu2020semantic}, we estimate the semantic shift of old prototypes without access to past samples. 

Suppose $\varphi_c^t$ denotes the prototype of category $c$ in session $t$ and $r$ is the learning session that the category belongs to. We have no access to the samples of category $c$ to update the prototype in session $t$ (when $t > r$). The semantic shift of class $c$ between two sessions can be represented as 
\begin{equation}
\Delta_c^{r\rightarrow t}= \varphi_c^t -\varphi_c^r,\quad 
    \varphi_c^r=\frac{1}{N_r^c}\sum_{n=1}^{N_r^c}\mathcal{F}(X_r^c,\theta_r).
\label{compute prototype}
\end{equation}
While we do not have access to data from the old class $c$, we can only estimate the shift of current task categories on old and new models. SDC \cite{yu2020semantic} used the  shift of current samples between two sessions $\delta_i^{t-1\to t} = e_i^t -e_i^{t-1}$ to estimate the update $\Delta_c^{t-1\rightarrow t}$.
However, computing the offset between each training sample and all prototypes of old classes in the current stage is computationally expensive. As the number of classes increases, the computational cost will gradually escalate. A more efficient approach is to directly compute the offset between the prototypes on old model and the prototypes of the new model which can be represented as
 \begin{equation}
 \begin{aligned}
  \varphi_i^{t-1\to t} = \varphi_i^t -\varphi_i^{t-1}
  \end{aligned}
  \label{eq 9}
 \end{equation}
We can compute $\varphi_i^{t-1}$ at the start of the current task with the model trained in task $t-1$. After training on the new task, we compute $\varphi_i^{t-1\to t}$ and use it to estimate $\Delta_c^{t-1\rightarrow t}$. So, we correct the semantic shift as
\begin{equation}
\begin{aligned}
     \widetilde{\Delta}_c^{t-1\to t} =& \frac{\sum \alpha_i \varphi_i^{t-1\to t}}{\sum \alpha_i}, c \notin C^t, \\
     \alpha_i =& \frac{\varphi_i^{t-1}\cdot\varphi_c^{t-1}}{||\varphi_i^{t-1}|| \times ||\varphi_c^{t-1}||}.
\label{eq 10}
\end{aligned}
\end{equation} 
Where, the offset weight $\alpha$ is replaced by the cosine similarity between old and new class prototypes. The semantic shift is replaced by the offset of the new class prototype on the old and new model. Overall, before retraining the classifier, we update the prototypes with
\begin{equation}
    \left\{
    \begin{array}{lc}
        \varphi_c = \varphi_c^{t-1} +  \widetilde{\Delta}_c^{t-1\to t} & , c \notin C^t \\
        \varphi_c = \frac{1}{N_c}\sum_i e_c& , c \in C^t, \\
    \end{array}
    \right.
\label{compte-mean}
\end{equation}
where $N_c$ denotes the number of images in class $c$. The detailed computational steps are listed in algorithm~\ref{alg:semantic_shift}.

\subsection{Unified classifier training }
  Previous work \cite{zhang2023slca,tang2023prompt,zhu2021prototype}  has attempted to retrain a unified classifier by modeling each class as a Gaussian distribution and sampling features from the distribution. We refer to this method as classifier alignment (CA) and adopt a similar approach that incorporates semantic shift estimation, which we denote as SSCA. Specifically,  we compute the class prototypes $P_c = \{\varphi_1,...,\varphi_C \}$ and covariance $\Sigma_c = \{\varsigma_1,...,\varsigma_C \}$ for each class after training process in each learning session. The calculation of class prototypes is based on Eq.~\ref{compte-mean}. Due to the capability of the trained backbone network to provide well-distributed representations, each class exhibits an unimodal distribution. Therefore, we form a normal distribution $\mathcal{N}(\mu_c, \Sigma_c)$ for each class with class prototype and variance. We sample features $\mathcal{V}_c=\{v_{c,1},...v_{c, S_n}\}$ from the distribution to obtain diverse samples, where $S_n$ is the number of the sample features for each class. Then, we use these features to train classification layers $\theta_{cls}$ with a commonly used cross-entropy loss as
\begin{equation}
    \mathcal{L}(\theta_{cls}, \mathcal{V}_c) = -\sum_{i=1}^{S_n*C} log \frac{\mathbf{e}^{(\theta_{cls}^j(v_i ))}}{\sum_{k\in C} \mathbf{e}^{(\theta_{cls}^k(v_i))}},
\label{retrain loss}
\end{equation}
where $C$ denotes all classes learned so far. We normalize the features and classifier the same as backbone training. 
\begin{algorithm}
\caption{Training Algorithm for SSIAT.}
\label{alg:Total}  
\hspace*{0.02in}{\bf Input:}
The total number of sessions $T$; The pre-trained ViT model with classifier $M_{\theta}$; The sequence data $D$.\\
\hspace*{0.02in}\vspace*{0.03in}{\bf Output:}
The model  $M_{\theta}$ after training in each learning session. 
\begin{algorithmic}[1]
\label{alg:CurB}
\STATE {Initialize the classifier $f_{\theta_{cls}}$ of the model $M_{\theta}$}
\STATE {\textbf{for} $t=1,...,T$ \textbf{do}}
\STATE \hspace*{0.2in}{\textbf{if} $t=1$ \textbf{then}}
\STATE \hspace*{0.4in}{Train model $M_{\theta}$ using $\mathcal{L}^{t}$ with $D_t$ to get $M_{\theta}^t$}
\STATE \hspace*{0.4in}{Compute the Prototypes of each class using formula~\ref{compute prototype}}
\STATE \hspace*{0.2in}{\textbf{else}}
\STATE \hspace*{0.4in}{Train model $M_{\theta}$ using $\mathcal{L}^{t}$ with $D_t$ to get $M_{\theta}^t$}
\STATE \hspace*{0.4in}{Compute the Prototypes of each new class using formula~\ref{compute prototype}}
\STATE \hspace*{0.4in}{Update old class Prototypes with the algorithm of Semantic Shift Estimation.}
\STATE \hspace*{0.4in}{Construct Gaussian distribution for each class with updated Prototypes and then sample feature samples $s_t$}
\STATE \hspace*{0.4in}{Retrain the whole classifier $f_{\theta_{cls}}$ using formula~\ref{retrain loss} with feature samples $s_t$}
\STATE {\textbf{end for}}
\end{algorithmic}
\end{algorithm}

In general, our training process consists of two stages for each incremental task. First, we fine-tune the shared adapter in the ViT backbone to update feature expression. Then, we employ the classifier retraining which utilizes the Gaussian distribution sampling of feature representations for each class. This process involves optimizing the prototype offset of the old class and utilizing sampling from class Gaussian distributions. The specific algorithm is described in algorithm~\ref{alg:Total}.

    


\begin{table*}[t]
\centering
\caption{Experimental results on four CIL benchmarks. All other methods are reproduced using the same seeds for a fair comparison.}
\scalebox{0.95}{
\begin{tabular}{ccccccccccc}
\toprule

 \multirow{2}*{Method} &  \multirow{2}*{Params} &\multicolumn{2}{c}{ImageNetR} &\multicolumn{2}{c}{ImageNetA} &\multicolumn{2}{c}{CUB200} &\multicolumn{2}{c}{CIFAR100}\\
&{}&$\mathcal{A}_{Last} \uparrow $&$\mathcal{A}_{Avg} \uparrow$&$\mathcal{A}_{Last} \uparrow $&$\mathcal{A}_{Avg} \uparrow$ &$\mathcal{A}_{Last} \uparrow$&$\mathcal{A}_{avg} \uparrow$ &$\mathcal{A}_{Last} \uparrow $&$\mathcal{A}_{avg} \uparrow$\\
\hline
Joint  & 86M & $81.72_{\pm 0.35}$	& -& $50.56_{\pm 1.75}$&	-&$88.17_{\pm 0.32}$ &	- & $89.71_{\pm 0.07}$& - \\
FT    & 86M&  $20.93_{\pm 0.86}$	&$40.35_{\pm 0.74}$& $6.03_{\pm 4.74}$&	$16.57_{\pm 5.8}$ & $22.05_{\pm 1.69}$  & $45.67_{\pm 2.04}$& $22.17_{\pm 1.09}$ &	$41.83_{\pm 1.60}$    \\ \midrule
SLCA \cite{zhang2023slca}        & 86M & $79.35_{\pm 0.28}$	&$83.29_{\pm 0.46}$& $61.05_{\pm 0.63}$&	$68.88_{\pm 2.31}$ &$84.68_{\pm 0.09}$&	$90.77_{\pm 0.79}$ &  $ 91.26_{\pm 0.37}$  & $94.29_{\pm 0.92}$                        \\
Adam-Adapter \cite{zhou2023revisiting} 
        &1.19M & $65.79_{\pm 0.98}$	&$72.42_{\pm 1.41}$& $48.81_{\pm 0.08}$&$58.84_{\pm 1.37}$ &$85.84_{\pm 0.08}$&	$91.33_{\pm 0.49}$   &$87.29_{\pm 0.27}$&	$91.21_{\pm 1.33}$  
    \\
Adam-SSF \cite{zhou2023revisiting}       & 0.2M& $66.61_{\pm 0.09}$	&$74.36_{\pm 1.00}$ 
& $48.94_{\pm 0.14}$	&$58.79_{\pm 2.82}$ & $85.67_{\pm 0.15}$	&$90.99_{\pm 0.76}$ &  $85.27_{\pm 0.21}$ &$89.90_{\pm 0.98}$  \\
Adam-VPT \cite{zhou2023revisiting}  & 0.04M&  
$65.29_{\pm 1.52}$	&$72.97_{\pm 0.56}$ & $29.29_{\pm 7.42}$& $39.14_{\pm 7.59}$ &  $85.28_{\pm 0.47}$& $90.89_{\pm 0.86}$ &$85.04_{\pm 1.04}$ & $89.49_{\pm 0.58}$\\
LAE \cite{gao2023unified}
  &0.19M &  
$72.29_{\pm 0.14}$	&$77.99_{\pm 0.46}$& $47.18_{\pm 1.17}$	&$58.15_{\pm 0.73}$ & $80.97_{\pm 0.51}$	&$87.22_{\pm 1.21}$&$85.25_{\pm 0.43}$ &$89.80_{\pm 1.20}$ \\
L2P \cite{wang2022learning}      & 0.04M &$72.34_{\pm 0.17}$& $77.36_{\pm 0.64}$ &$44.04_{\pm 0.93}$&$51.24_{\pm 2.26}$ &$67.02_{\pm 1.90}$ &$79.62_{\pm 1.60}$&	   $84.06_{\pm 0.88}$ &$88.26_{\pm 1.34}$\\
ADA \cite{ermis2022memory}      & 1.19M &$73.76_{\pm 0.27}$& $79.57_{\pm 0.84}$ &$50.16_{\pm 0.20}$&$59.43_{\pm 2.20}$ &$76.13_{\pm 0.94}$ &$85.74_{\pm 0.26}$&	   $88.25_{\pm 0.26}$ &$91.85_{\pm 1.32}$\\
DualPrompt \cite{wang2022dualprompt}    & 0.25M & $69.10_{\pm 0.62}$ &  $74.28_{\pm 0.66}$ &$53.19_{\pm 0.74}$ &  $64.59_{\pm 0.08}$ &  $68.48_{\pm 0.47}$ &$80.59_{\pm 1.50}$ &  $86.93_{\pm 0.24}$ &$91.13_{\pm 0.32}$  \\
CODAPrompt \cite{smith2023coda}    & 3.84M & $73.31_{\pm 0.50}$ &  $78.47_{\pm 0.53}$ &$52.08_{\pm 0.12}$ &  $63.92_{\pm 0.12}$ &  $77.23_{\pm 1.12}$ &$81.90_{\pm 0.85}$  &  $83.21_{\pm 3.39}$ &$87.71_{\pm 3.17}$  \\
RANPAC\cite{mcdonnell2024ranpac} & 3.84M & $77.83_{\pm 0.23}$ &  $82.97_{\pm 0.33}$ &$60.46_{\pm 0.89}$ &  $68.11_{\pm 2.95}$ &  $89.95_{\pm 0.65}$ &$93.67_{\pm 0.59}$  &  $91.58_{\pm 0.22}$ &$94.44_{\pm 0.91}$\\
EASE \cite{zhou2024expandable}   & 11.9M&   $76.04_{\pm 0.16}$ &  $81.38_{\pm 0.36}$ &$55.15_{\pm 0.78}$ &  $64.15_{\pm 1.47}$ & $85.21_{\pm 0.50}$ &$90.92_{\pm 0.76}$ &  $88.25_{\pm 0.54}$ & $92.02_{\pm 1.04}$ \\ 
SSIAT\cite{tan2024semantically} & 1.19M& $79.38_{\pm 0.59}$ &  $83.63_{\pm 0.43}$ &$62.43_{\pm 1.63}$ &  $70.83_{\pm 1.63}$ &  $88.75_{\pm 0.38}$ &$93.00_{\pm 0.90}$ &  $91.35_{\pm 0.26}$ & $94.35_{\pm 0.60}$ \\
\textbf{Ours}   & 1.19M& $79.61_{\pm 0.40}$ &  $83.71_{\pm 0.42}$ &$62.73_{\pm 1.01}$ &  $71.05_{\pm 1.73}$ &  $88.32_{\pm 0.48}$ &$92.95_{\pm 0.55}$ &  $91.51_{\pm 0.20}$ & $94.29_{\pm 0.88}$   \\ 
\hline
\end{tabular}}

\label{performance}
\end{table*}

\begin{figure*}[htbp]
\centering
\includegraphics[scale=0.60]{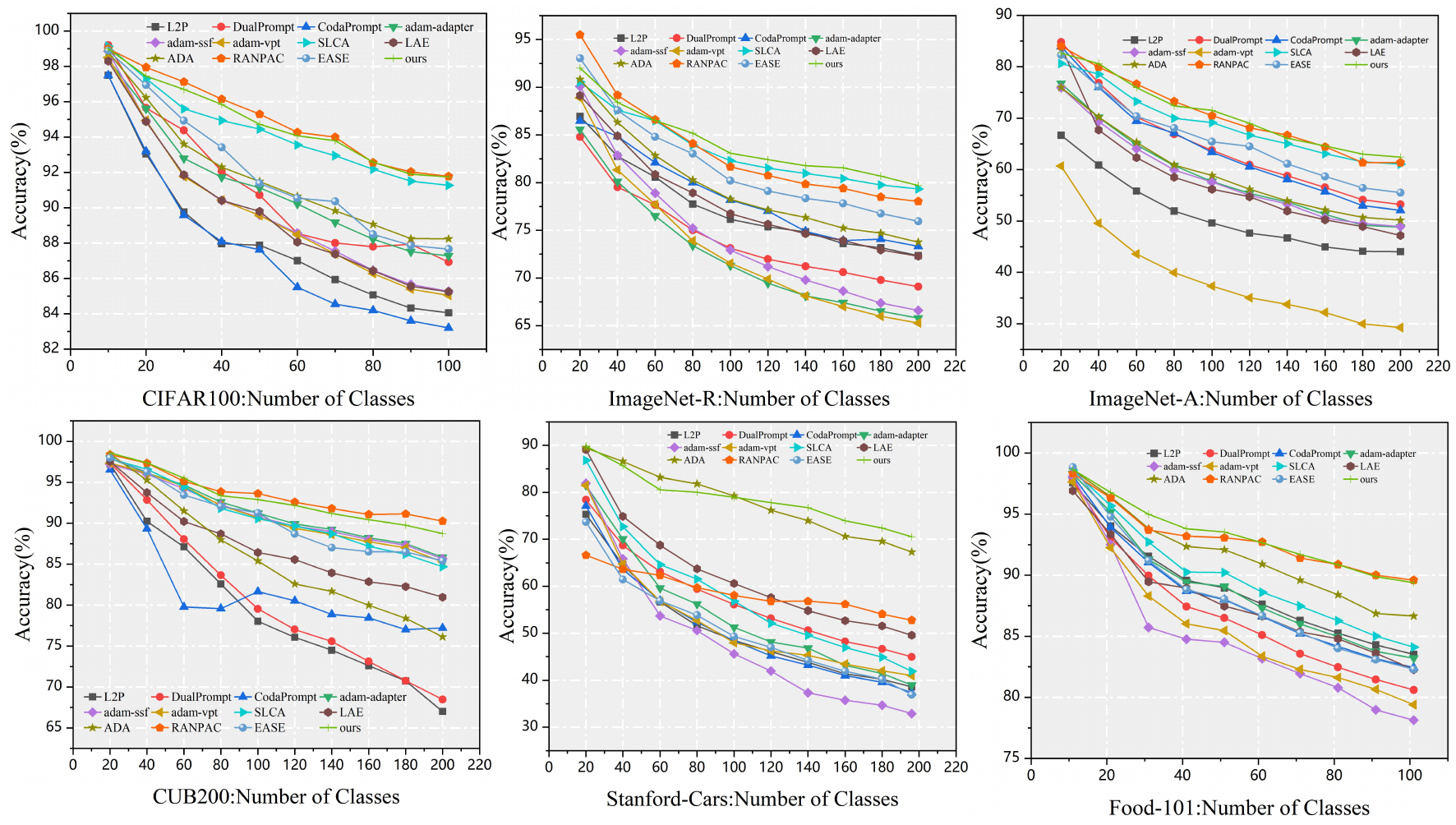}    
\caption{The performance of each learning session on six CIL datasets. (a) ImageNetR; (b) ImageNetA; (c) CUB200; (d) CIFAR100; (e) Stanford Cars; (f) Food-101 . These curves are plotted by calculating the average performance across three different seeds for each incremental session. }
\label{fig5}
\end{figure*}

\section{Experiments}
\label{Exp}
\subsection{Datasets and Evaluation Protocols}
\textbf{Dataset:} We evaluate our method on six commonly used CIL benchmarks and one cross-domain CIL dataset. We randomly split the dataset into 10 or 20 learning sessions in the CIL setting. Furthermore, we evaluate our method on few-shot class-incremental learning and hierarchical class-incremental learning datasets. 
CIFAR100 \cite{krizhevsky2009learning} is a widely used dataset in CIL which consists of 60000 images, belonging to 100 different categories. CUB200 \cite{wah2011caltech} is a dataset that contains approximately 11,788 images of 200 bird species with fine-grained class labels.  Additionally, we also follow recent work \cite{zhou2023revisiting, zhang2023slca} to use the other three datasets which have a large domain gap with pre-training data. ImageNetR \cite{hendrycks2021many} consists of 30,000 images with 200 categories. Although its categories overlap with ImageNet-21K \cite{russakovsky2015imagenet}, the images belong to a different domain. ImageNetA \cite{hendrycks2021natural} is a real-world dataset that consists of 200 categories. This dataset exhibits significant class imbalance, with some categories having only a few training samples. Stanford Cars \cite{krause20133d} is a fine-grained car dataset that consists of 196 categories including a total of 16185 images, and the training and test sets are divided in a 1:1 ratio. Food-101 \cite{bossard2014food} is a dataset containing 101 fine-grained foods with a total of 101k images. These two datasets are fine-grained datasets, which have a large domain gap with the pre-training data. VTAB \cite{zhai2019large} is a complex dataset that consists of 19 tasks covering a broad spectrum of domains and semantics. We follow previous work \cite{zhou2023revisiting} to select 5 tasks to construct a cross-domain CIL dataset.

\begin{table}[t]
\centering
\caption{Experimental results for long-sequences (20 incremental sessions) on ImageNetR and ImageNetA dataset. }
\scalebox{0.82}{
\begin{tabular}{cccccc}
\toprule
 \multirow{2}*{Method} &\multicolumn{2}{c}{ImageNetR} &\multicolumn{2}{c}{ImageNetA}  \\
 & $\mathcal{A}_{Last}\uparrow$& $\mathcal{A}_{Avg}\uparrow$&$\mathcal{A}_{Last}\uparrow$ &$\mathcal{A}_{Avg}\uparrow$ \\
\hline
SLCA \cite{zhang2023slca}   & $74.63_{\pm 1.55}$& $79.92_{\pm 1.29} $ &  $36.69_{\pm 21.31}$ & $56.35_ {\pm 7.09}$
              \\
Adam-adapter\cite{zhou2023revisiting}    &$57.42_{\pm 0.84}$  & $64.75_{\pm 0.79}$  &  $48.65_{\pm 0.12}$ & $59.55_{\pm 1.07} $            \\
Adam-ssf\cite{zhou2023revisiting} & $64.30_{\pm 0.94}$ & $72.42_{\pm 1.47}$ &$47.27_{\pm 4.34}$& $58.36_{\pm 4.70}$                   \\
Adam-prompt\cite{zhou2023revisiting} & $59.90_{\pm 1.13}$ & $68.02_{\pm 1.02}$ &$29.93_{\pm 4.88}$& $39.13_{\pm 4.19}$                \\
LAE \cite{gao2023unified}     & $69.86_{\pm0.43}$  & $77.38_{\pm 0.61}$ & $39.52_{\pm 0.78}$ & $ 51.75_{\pm 2.15} $\\
L2P \cite{wang2022learning}   & $69.64_{\pm 0.42}$	& $75.28_{\pm 0.57}$ & $40.48_{\pm 1.78}$ & $49.62_{\pm 1.46}$ &\\
DualPrompt \cite{wang2022dualprompt} & $66.61_{\pm 0.58}$ & $72.45_{\pm 0.37}$ & $42.28_{\pm 1.94}$ & $53.39_{\pm 1.64}$ & \\
CODAPrompt \cite{smith2023coda} & $69.96_{\pm 0.50}$ & $75.34_{\pm 0.85}$ & $44.62_{\pm 1.92}$ & $54.86_{\pm 0.50}$ & \\ 
EASE \cite{zhou2024expandable} & $73.82_{\pm 0.55}$ & $80.31_{\pm 0.70}$ & $44.41_{\pm 2.65}$ & $56.23_{\pm 3.46}$ & \\
\textbf{SSIAT (Ours)} & $75.67_{\pm 0.14}$ & $82.30_{\pm 0.36}$ & $58.92_{\pm 1.82}$ & $67.73_{\pm 0.25}$ \\ 
\bottomrule
\end{tabular}}

\label{long}
\end{table}

\begin{table}[t]
\centering
\caption{Experimental results for fine-grained dataset on the CIL setting. }
\scalebox{0.82}{
\begin{tabular}{cccccc}
\toprule
 \multirow{2}*{Method} &\multicolumn{2}{c}{Food101} &\multicolumn{2}{c}{StanfordCars}  \\
 & $\mathcal{A}_{Last}\uparrow$& $\mathcal{A}_{Avg}\uparrow$&$\mathcal{A}_{Last}\uparrow$ &$\mathcal{A}_{Avg}\uparrow$ \\
\hline
SLCA \cite{zhang2023slca}   & $86.66_{\pm 0.22}$& $91.57_{\pm 0.36} $ &  $49.58_{\pm 1.93}$ & $62.32_ {\pm 1.37}$
              \\
Adam-adapter\cite{zhou2023revisiting}    &$83.51_{\pm 0.25}$  & $88.86_{\pm 0.42}$  &  $38.60_{\pm 0.15}$ & $50.66_{\pm 0.59} $            \\
Adam-ssf\cite{zhou2023revisiting} & $80.60_{\pm 0.32}$ & $86.79_{\pm 0.40}$ &$44.96_{\pm 0.62}$& $56.94_{\pm 0.65}$                   \\
Adam-prompt\cite{zhou2023revisiting} & $82.44_{\pm 0.92}$ & $88.14_{\pm 0.25}$ &$37.37_{\pm 5.95}$& $50.47_{\pm 5.58}$                \\
LAE \cite{gao2023unified}     & $82.26_{\pm 0.49}$  & $87.88_{\pm 1.94}$ & $41.87_{\pm 1.55}$ & $ 57.78_{\pm 1.50} $\\
L2P \cite{wang2022learning}   & $79.40_{\pm 1.41}$	& $85.70_{\pm 0.86}$ & $32.87_{\pm 1.09}$ & $48.02_{\pm 2.25}$ &\\
DualPrompt \cite{wang2022dualprompt} & $78.13_{\pm 0.12}$ & $84.86_{\pm 0.96}$ & $32.87_{\pm 1.09}$ & $48.02_{\pm 2.25}$ & \\
CODAPrompt \cite{smith2023coda} & $83.21_{\pm 0.10}$ & $88.85_{\pm 0.40}$ & $39.01_{\pm 3.27}$ & $53.71_{\pm 1.79}$  & \\ 
EASE \cite{zhou2024expandable} & $81.99_{\pm 0.32}$ & $87.95_{\pm 0.61}$ & $36.46_{\pm 1.02}$ & $62.32_{\pm 1.37}$  &\\
PANPAC \cite{zhou2024expandable} & $89.02_{\pm 0.99}$ & $92.78_{\pm 0.23}$ & $52.79_{\pm 39.54}$ & $55.54_{\pm 39.17}$\\
\textbf{SSIAT (Ours)} & $89.38_{\pm 0.11}$ & $93.22_{\pm 0.40}$ & $67.43_{\pm 1.47}$ & $75.53_{\pm 1.26}$ \\ 
\bottomrule
\end{tabular}}

\label{Fine-grained}
\end{table}

\textbf{Implementation details:}
We use ViT-B/16 \cite{dosovitskiy2020image} as the pre-trained model, which is pre-trained on ImageNet-21K \cite{russakovsky2015imagenet}. The initial learning rate is set as 0.01 and we use the cosine Anneal scheduler. In our experiments, we train the first session for 20 epochs and 10 epochs for later sessions.  Following previous papers \cite{zhou2023revisiting,zhang2023slca}, we use common evaluation metrics in CIL. Specifically, we report the last session accuracy $\mathcal{A}_{Last}$ and average accuracy of the whole incremental sessions $\mathcal{A}_{Avg}=\frac{1}{T} \sum_{i=1}^T \mathcal{A}_i $. \textit{We utilize three different seeds to generate three different class orders for evaluating various methods. We report the mean and standard deviation based on the three experiments.} See codes\footnote{\url{https://github.com/HAIV-Lab/SSIAT}} for more details.

\subsection{Experiment Results}
For a fair comparison, we compare our methods with SOTA CIL methods based on the pre-trained vision transformer model. We compare our methods with prompt-based methods L2P \cite{zhou2022learning}, DualPrompt \cite{wang2022dualprompt}, CODAPrompt \cite{smith2023coda}, fine-tuning methods SLCA \cite{zhang2023slca}, and adapter-based method \cite{zhou2023revisiting,gao2023unified,ermis2022memory,zhou2024expandable}. Joint training and Fine-tuning are regarded as lower and upper bound in the CIL setting. Tab.~\ref{performance} shows $\mathcal{A}_{Avg}$ and $\mathcal{A}_{Last}$ with three different seeds on four CIL benchmarks.

\textbf{CUB200 $\&$ CIFAR100:} We first report the results of each method on the CUB200 and CIFAR100 datasets. Since these two datasets overlap with the pre-training data, methods based on a pre-trained model achieve a huge improvement in performance compared with methods that are trained from scratch. For example, as shown in Tab.~\ref{performance}, the average accuracy on L2P, DualPrompt, and CODAPrompt reached 88.26\%, 91.13\%, and 87.71\% on CIFAR100, respectively. Nevertheless, our method still outperforms those prompt-based methods. Besides, our method does not require the construction of a prompt pool which allows each task to learn specific prompt parameters. The adapter is shared across tasks and our method avoids the parameter expansion with tasks increasing.  Even though the Adam-adapter/SSF/prompt only needs to train in the first stage which requires less training time, the performance of those methods is inferior to our proposed method.  Although the performance of SLCA is comparable to our method in CIFAR100, the number of tuning parameters of our method is much smaller. Besides that, the average performance of our method on CUB200 is 93.00$\%$, nearly 2.3$\%$  improvement over SLCA. Fig.~\ref{fig5} (c) (d) shows the incremental accuracy of each session on CUB200 and CIFAR100 and our method is always at the top of all lines in the incremental process.

\begin{table}[t]
\centering
\caption{Experimental results for different methods on VTAB dataset which contain 5 datasets from different domains.}
\scalebox{0.90}{
\begin{tabular}{ccccccl}
\toprule
Method & Sess.1    & Sess.2    & Sess.3    & Sess.4  & Sess.5   & Avg$\uparrow$       \\
\hline
 Adam-adapter\cite{zhou2023revisiting} &  87.60 & 86.07 & 89.14 & 82.72 & 84.35 & 85.97
 \\
Adam-ssf\cite{zhou2023revisiting} & 89.60& 88.21 & 89.94 & 80.50 & 82.38 & 86.13 \\
Adam-vpt\cite{zhou2023revisiting}& 90.20 & 87.57 & 89.69 & 80.39 & 82.18 & 86.01  \\
SLCA\cite{zhang2023slca}  &  94.80 & 92.43 & 93.54 & \textbf{93.98} & 94.33 & 93.82               \\
LAE  \cite{gao2023unified}  & \textbf{97.99} &  85.26 & 79.68 & 78.78 & 74.36 & 83.21                  \\
EASE \cite{zhou2024expandable}& 96.00 &92.43&92.66 & 92.76 &93.43 &93.46\\
\textbf{SSIAT (Ours) }&96.40&\textbf{93.14}&\textbf{93.96}&93.86&\textbf{94.38}&\textbf{94.34}
\\
\bottomrule
\end{tabular}}
\label{VTAB}
\end{table}

\begin{table}[h]
\centering
\caption{
The experiment results of hierarchical class-incremental learning on the CIFAR dataset.
}
\begin{tabular}{cccc}
\toprule
Method           & Hier.1          & Hier.2   & Avg$\uparrow$   \\ \hline
LwF \cite{li2017learning}              & 62.44$\rightarrow$ 76.44 & 74.45 & 75.44 \\
iCaRL   \cite{rebuffi2017icarl}         & 68.01$\rightarrow$ 76.21 & 71.60  & 73.90  \\
Adam-adapter \cite{zhou2023revisiting}     & 94.95$\rightarrow$ 91.66 & 90.47 & 91.06 \\
Adam-ssf   \cite{zhou2023revisiting}      & 95.02$\rightarrow$ 92.40 & 90.13 & 91.26 \\
Adam-vpt-shallow \cite{zhou2023revisiting}& 88.92$\rightarrow$ 88.92 & 75.00   & 81.96 \\
Adam-vpt-deep  \cite{zhou2023revisiting}  & 93.48$\rightarrow$ 89.03 & 90.82 & 89.92 \\
\textbf{SSIAT(Ours) }          & 94.96$\rightarrow$ 94.41 & 92.24 & 93.32 \\
\bottomrule
\end{tabular}
\label{hier}
\end{table}

\begin{figure}[t]
\centering
\includegraphics[scale=0.53]{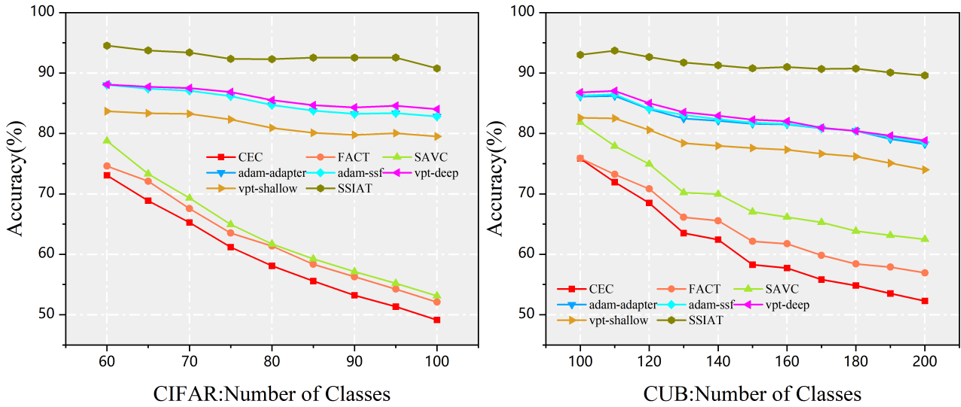}        
\caption{Experimental results of few-shot incremental learning on cifar100 and cub200 datasets.}
\label{few-shot}
\centering
\end{figure}

\begin{table*}[htb!]
\centering
\caption{Experimental results for baselines with different efficient tuning methods on CIFAR100. We report the overall performance of each session and the average performance. }
\begin{tabular}{cccccccccccll}
\toprule
PET Method &  Params & Sess.1                  & Sess.2                   & Sess.3                & Sess.4                  & Sess.5                    & Sess.6                 & Sess.7                 & Sess.8                  & Sess.9                  & Sess.10  & \multicolumn{1}{c}{Avg$\uparrow$}      \\ 
\midrule \midrule
SSF   \cite{lian2022scaling}         & 0.2M                             & \textbf{98.50}                & 91.90                & 88.57                & 85.02                & 83.92                & 78.70                & 77.79                & 77.89                & 73.02                & \multicolumn{1}{c}{74.91} & \multicolumn{1}{c}{83.03}                        \\

VPT-deep \cite{jia2022visual}      &       0.046M                       & 97.60                 & 69.15                 & 68.70                & 56.60                & 55.56                & 48.87                & 55.97                & 56.05                & 53.48                &   55.21                        &   61.72                                            \\
VPT-shallow \cite{jia2022visual}   &      0.004M                        & 98.40 & 92.95 & 88.80 & 92.06 & 87.26 & 86.37 & 85.64 & 85.31 & 85.36 &    85.10                       &     88.72                                          \\
\textbf{Adapter} \cite{chen2022adaptformer}     &       1.19M                       & \textbf{98.50}               & \textbf{95.35}                & \textbf{91.60}               & \textbf{91.08}                 & \textbf{90.92}                & \textbf{90.08}            & \multicolumn{1}{l}{\textbf{89.80}} & \multicolumn{1}{l}{\textbf{89.62}} & \multicolumn{1}{l}{\textbf{88.98}} &  \textbf{89.29}                       &  \textbf{91.52}                                                \\
\bottomrule
\end{tabular}
\label{tuning}
\end{table*}


\textbf{ImageNetR $\&$ ImageNetA:} We report the performance on ImageNetR and ImageNetA in Tab.~\ref{performance}. These two datasets are more difficult due to the domain gap with the pre-training data. It can be seen that the performance of each method on these two datasets is lower than CIFAR100 and CUB200. Besides, we can see that SLCA outperforms other previous methods significantly on these two datasets. Notably, SLCA achieves an impressive last accuracy on ImageNetR, surpassing the other methods. In contrast, our method achieves SOTA-level performance on both datasets with fewer tuning parameters. Based on Fig.~\ref{fig5}, the performance of our method is slightly higher than SLCA in several learning sessions with fewer tuning parameters on the ImageNetR dataset. On the ImageNetA dataset, our method achieves the last accuracy of 62.43\%, surpassing SLCA by 1.39\%. The average accuracy across all sessions is 70.83\%, showing a 2\% improvement. 

Additionally, we evaluate the performance of each method under the condition of long sequences. In this setting, each session consists of only 10 classes, and the results are summarized in Tab.~\ref{long}. Our method also maintains excellent performance in terms of $\mathcal{A}_{Last}$ and $\mathcal{A}_{Avg}$. The performance of SLCA is highly dependent on the class order in which the training data appears, resulting in a substantial variance in $\mathcal{A}_{Last}$ on ImageNetA. In contrast, the Adam-based methods remain relatively stable in long-sequence settings. For Adam-SSF, the long sequence only leads to a nearly $2\%$ performance drop in ImageNetR. However, for SLCA, its performance drops by $5\%$ on ImageNetR and nearly $10\%$ on ImageNetA.  In comparison, our method demonstrates excellent stability on long sequences and outperforms other methods by a large margin.

\textbf{Food101 $\&$ StanfordCars:} We report the results of Food101 and StanfordCars on Tab.~\ref{Fine-grained}. These two datasets are fine-grained datasets, and it is challenging for models to achieve good recognition performance solely based on pre-trained knowledge on such datasets. It can be observed that both first session adaptation-based methods and prompt-based methods perform poorly on these two datasets. In contrast, methods retraining classifiers, such as SLCA and our proposed method, demonstrate significant improvements in recognition performance compared to these methods. Besides, our method addresses the semantic shift of class prototypes, resulting in improved performance on both datasets.

\textbf{Comparison to traditional CIL methods:} We conduct evaluations by comparing our approach to SOTA traditional CIL methods shown in Tab.~\ref{traditional}. We replace the Resnet backbone with the pre-trained ViT model for fair comparison. The results indicate that the performance of iCaRL tends to be inferior compared to SOTA model expansion methods and our proposed method, even when past samples are stored. It can be observed that methods such as Foster and Der, which dynamically expand feature extraction networks, achieve impressive results on ImageNetR. The average accuracy of these methods is only $2\%$ lower than our method. However, on ImageNetA, where there are few-shot samples for many classes, these methods exhibit low performance.

\begin{table*}[htb!]
\centering
\caption{Ablation results for unified classifier training and semantic shift estimation on ImageNetA. We report the overall performance of each session and the average performance. We run the experiments with three seeds and reported the average performance.}
\begin{tabular}{ccccccccccccl}
\toprule
Classifier              & Method & Sess.1    & Sess.2    & Sess.3   & Sess.4    & Sess.5   & Sess.6   & Sess.7  & Sess.8   & Sess.9   & Sess.10   & Avg$\uparrow$     \\\midrule \midrule
\multirow{3}{*}{Linear} & w/o CA     & 74.65  & 68.37 & 63.90 & 58.82 & 58.02 & 55.48 & 54.03 & 52.89 & 51.62 & 52.13 &  58.99                    \\
 & w/ CA     & 74.65 & 71.59 & 67.93 & 64.24 & 62.08 & 60.90 & 59.03 & 57.32 & 56.41 & 56.85 & 63.10                     \\
 & w/ SSCA    & 74.65 & 70.92 & 67.64 & 63.91 & 62.65 & 60.96 & 60.38 & 58.55 & 58.13 & 57.77 & 63.55                        \\
                \hline
\multirow{3}{*}{Cosine} & w/o CA    & 82.66&	77.78 &	72.20 &	67.63 &	66.01 &	63.18 &59.97 &	59.35&	58.93 &	57.91 &	66.56
        \\

& w/ CA &   82.66 &	79.70 &	74.56 &	70.40	& 68.19 &	65.66	& 63.40 &	61.77 & 60.70 &	59.78 &	68.68
                \\
& w/ SSCA  & 82.66 &	80.84 & 72.94 &	71.26 &	68.97	& 66.78 &	63.69 &	63.64 &63.12 &	61.62 &	69.55
            \\
\bottomrule
\end{tabular}
\label{classifier}
\end{table*}

\begin{table}[t]
\centering
\caption{Linear probing results of different training ways on four datasets. We retrain the classifier using all the data on the fixed-trained backbone. }
\begin{tabular}{ccccl}
\toprule
Method & CIFAR100   & ImageNetR   & ImageNetA   &  CUB200 \\
\hline
 No-Adapt.&  86.08 & 68.42 & 33.71 &86.77
 \\
   First-Adapt.&  91.33 &78.02 &63.53 & 89.27\\
   All-Adapt.& \textbf{92.57 }& \textbf{82.02} &\textbf{65.96} &\textbf{89.86} \\
   $\Delta \uparrow$ & $ 1.24\%$ &$ 4.00\%$ & $ 2.43\%$ & $0.59\%$

\\
   
\bottomrule
\end{tabular}

\label{first}
\end{table}

\begin{table}[t]
\centering
\caption{Experimental results of different adapter structures. We report the average performance and standard deviation.}
\scalebox{0.95}{
\begin{tabular}{ccccl}
\toprule
Structure & Params &CIFAR   & ImageNetR  & ImageNetA   \\
\hline
 AdaptMLP-P \cite{chen2022adaptformer}  &1.19M  & $94.35_{\pm 0.60}$&$83.63_{\pm 0.43}$  & $70.83_{\pm 1.63}$
 \\
  AdaptMLP-S \cite{chen2022adaptformer} & 1.19M  &  $94.16_{\pm 0.88}$&$83.19_{\pm 0.47}$ & $71.00_{\pm 1.52}$\\
   Convpass \cite{jie2022convolutional} & 1.63M &$94.08_{\pm 0.99}$ & $83.64_{\pm 0.35}$ & $69.96_{\pm 1.09}$ \\
    Adapter \cite{houlsby2019parameter} & 2.38M  &  $94.26_{\pm 0.91}$& $83.65_{\pm 0.50}$&  $70.94_{\pm 1.42}$ \\
\bottomrule
\end{tabular}}
\label{adapter-type}
\end{table}

\begin{figure}[t]
\centering
\includegraphics[scale=0.29]{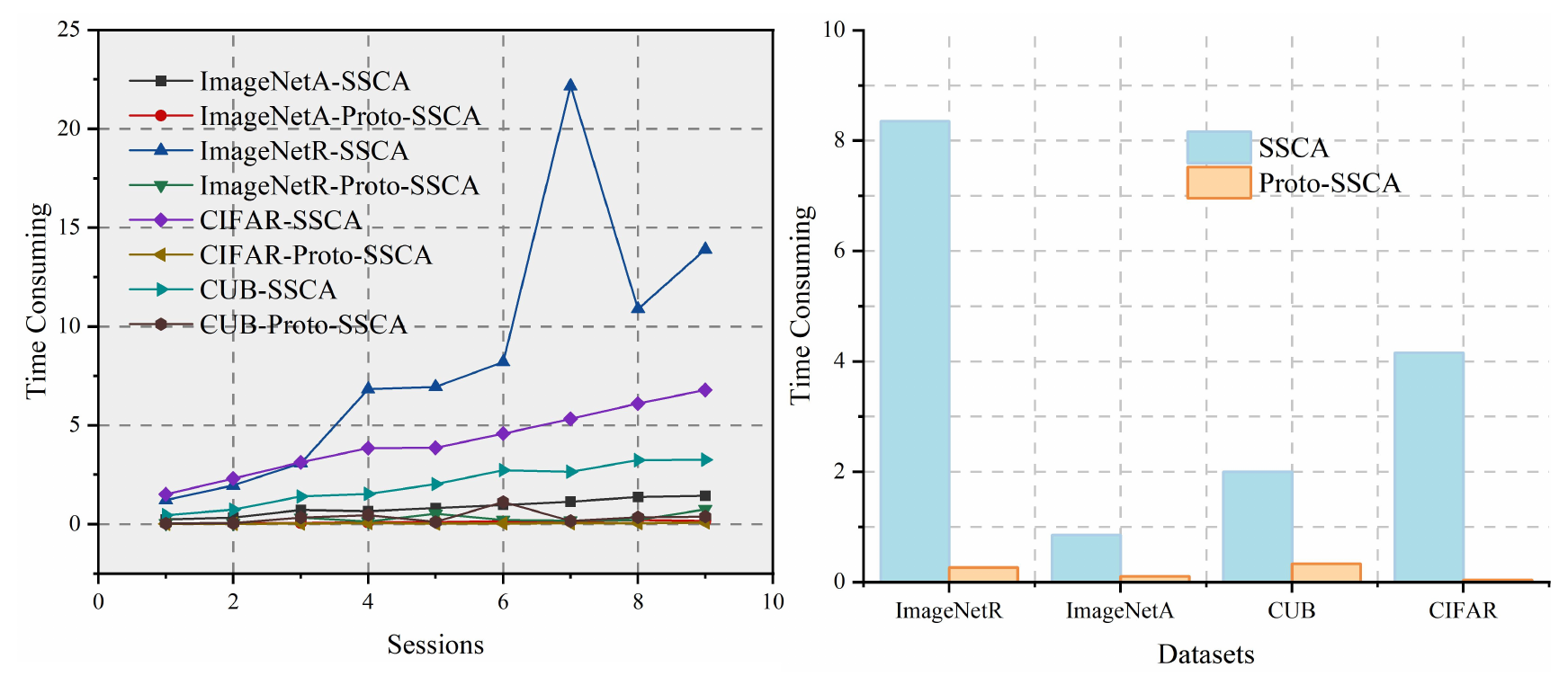}        
\caption{The comparison of the time costs between prototype-based semantic shift and sample-based semantic shift. The left shows the time consumed by different datasets and the right shows the time consumed by different sessions.}
\label{time cost}
\centering
\end{figure}

\subsection{More CIL Settings and Results}

\textbf{Cross-Domain Class-Incremental Learning.}   VTAB is a cross-domain CIL dataset where each task provides training data from a different domain. Based on the results presented in Tab.~\ref{VTAB}, it can be observed that both SLCA and our method perform well in cross-domain CIL. Specifically, in the last incremental stage, our method achieves an accuracy that is $12\% $ higher than the Adam-based methods. Adam-based methods only perform fine-tuning in the first task and are not able to adapt well to subsequent tasks on the cross-domain dataset. Compared to Adam-based methods, EASE exhibits excellent performance while needing more model storage for expanded adapters.

\textbf{Few-Shot Class-Incremental Learning.} We report the performance of our method on CIFAR100 and CUB200 datasets in the few-shot incremental learning (FSCIL) setting, shown in Fig.~\ref{few-shot}. In the few-shot class-incremental learning setting, the model in the first stage learns a significantly larger number of classes compared to the incremental sessions that follow, and in the subsequent sessions, each class is provided with only $5$ training samples. We run the first session adaptation-based methods and some SOTA methods in few-shot incremental learning. It can be seen that the methods using the pre-trained  ViT network have significantly improved compared to other methods. Furthermore, our method shows clear advantages throughout the entire incremental process.

\textbf{Hierarchical Class-Incremental Learning.} We report the performance of our method on the CIFAR100 dataset in hierarchical class-incremental learning (HCIL) setting, shown in Fig.~\ref{hier}. In contrast to the traditional setting of treating all classes equally in the incremental process, HIL requires that any new classes learned in new sessions are a subclass from the last session. After the entire incremental process, the classes learned form a hierarchical tree structure. The model need to provide correct classification at each level of the hierarchy. It can be observed that compared to the first session adaptation method and traditional incremental learning approaches, our method achieves higher performance at all levels of the hierarchical structure. 

\subsection{Ablation Study}
\textbf{Baselines with different PET methods:} Tab.~\ref{tuning} shows the results of baselines with three different parameter-efficient tuning methods in each incremental session. It can be observed that the pre-trained model with an adapter achieves the best performance in terms of both the last session accuracy and average accuracy. Fig.~\ref{fig1} demonstrates that tuning with an adapter achieves a better balance between learning new classes and retaining knowledge of old classes. Both VPT-deep and SSF methods tend to prioritize learning new categories, which leads to increased forgetting of previously learned categories. Although VPT-shallow performs well on CIFAR, its limited parameters hinder the model from incrementally learning new classes on ImageNetR. More results on the other datasets can be found in the \textit{Supp}.

\textbf{Unified classifier retraining vs. Separate local classifier:} As we train separate task-specific classifiers in each incremental session, we propose to retrain the classifier to find the optimal decision boundary for all the classes.  Tab.~\ref{classifier} displays the ablation experiments of the classifier re-trained on ImageNetA which is the most difficult benchmark. It can be observed that whether it is a linear or a cosine classifier, retraining the classifier leads to a significant performance improvement.  Additionally, incorporating the computation of prototype semantic shifts further enhances the performance by an additional $2\%$ in the cosine classifier. Compared to the classifier alignment methods that do not involve computing updated prototypes, our method demonstrates its superiority as the incremental stages progress. More results on the other datasets can be found in the \textit{Supp}.

\textbf{Progressively tuning vs. first session adaptation:} Tab. \ref{first} shows the linear probing results of different adaption ways. After finishing the training of the last session, we freeze the pre-trained backbone and only train the classifier using all the samples. It is evident that not performing tuning and solely freezing the pre-trained model leads to the worst performance, regardless of the dataset. First-session adaptation proves to be a good choice as it reduces training time and works well for datasets like CIFAR100 and CUB200. However, for datasets such as ImageNetA and ImageNetR, which have significant domain gaps from the pre-trained model, relying solely on first-session adaptation is suboptimal. By continuously fine-tuning the adapter, we observe that the backbone exhibits stronger representability compared to only tuning in the first session. 

\textbf{Different structures of the adapter:} In this paper, we follow AdaptFormer \cite{chen2022adaptformer} to use parallel adapterMLP as the adapter structure. We also delve deeper into different adapter structures such as Adapter \cite{houlsby2019parameter} and Convpass \cite{jie2022convolutional}. Although these different tuning structures may exhibit performance differences under static settings, the performance differences among those adapter structures are minimal in the context of CIL shown in Tab.~\ref{adapter-type}. This offers us the flexibility to employ various adapter structures within the context of the CIL paradigm.

\textbf{Runtime cost of semantic shift.} Performing semantic shift at each incremental session incurs extra computation time, and as the number of classes grows, the time cost gradually increases. Therefore, it is necessary to discuss the time costs associated with the semantic shift computation. We report the results of the time cost of the prototype-based method and sample-based method in each session shown in Fig.~\ref{time cost}. It can be seen from the right figure that as the number of sessions increases, the time cost associated with the semantic shift continues to grow. Besides, it can be seen that the sample-based semantic shift method incurs significantly higher time costs compared to the prototype-based method across various datasets.


\begin{table}[t]
\centering
\caption{Comparison to traditional CIL methods on ImageNetR and ImageNetA dataset.  }
\scalebox{0.90}{
\begin{tabular}{cccccc}
\toprule
 \multirow{2}*{Method} &\multicolumn{2}{c}{ImageNetR} &\multicolumn{2}{c}{ImageNetA}  \\
 & $\mathcal{A}_{Last}\uparrow$& $\mathcal{A}_{Avg}\uparrow$&$\mathcal{A}_{Last}\uparrow$ &$\mathcal{A}_{Avg}\uparrow$ \\
\hline
iCaRL \cite{rebuffi2017icarl}   & $61.70_{\pm 0.56}$  & $71.34_{\pm 0.67}$   & $29.32_{\pm 2.36}$ & $40.11_{\pm 1.36}$
              \\
Foster \cite{wang2022foster}    & $75.87_{\pm 0.38}$  & $81.54_{\pm 0.82}$  & $12.44_{\pm 17.45}$   & $17.01_{\pm 20.44}$           \\
Der \cite{yan2021dynamically}   & $75.63_{\pm 0.86}$ & $81.13_{\pm 0.11}$ & $38.43_{\pm 2.39}$  &  $46.43_{\pm 3.29}$                 \\
Memo \cite{zhou2022model} & $65.38_{\pm 0.90}$ & $73.80_{\pm 0.86} $& $28.45_{\pm 2.37}$  & $40.27_{\pm 1.22}$               \\

\textbf{SSIAT (Ours)} &$79.61_{\pm 0.40}$ &  $83.71_{\pm 0.42}$ & $62.73_{\pm 1.01}$ &  $71.05_{\pm 1.73}$\\ 
\bottomrule
\end{tabular}
}
\label{traditional}
\end{table}

\begin{figure*}[t]
\centering
\includegraphics[scale=0.5]{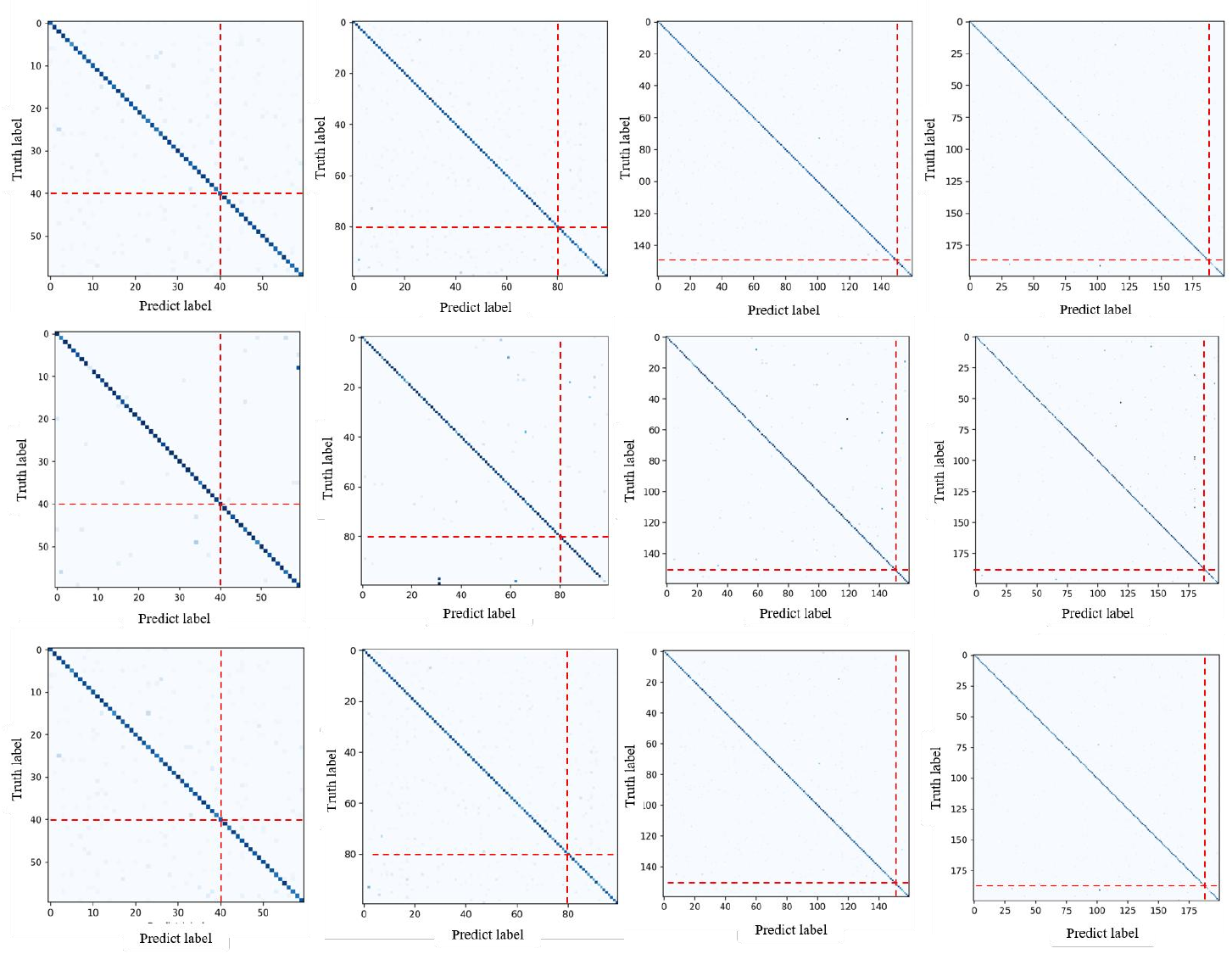}        
\caption{The confusion matrix of our proposed method on session 2,4,7,9 on ImageNetA, CUB and ImageNetR datasets. The bottom-right quadrant of the intersection points of the decision boundaries depicts the novel classes being learned in the current session.  }
\label{confusion_matrix}
\centering
\end{figure*}

\subsection{More Analysis}
\begin{figure}[t]
\centering
\includegraphics[scale=0.48]{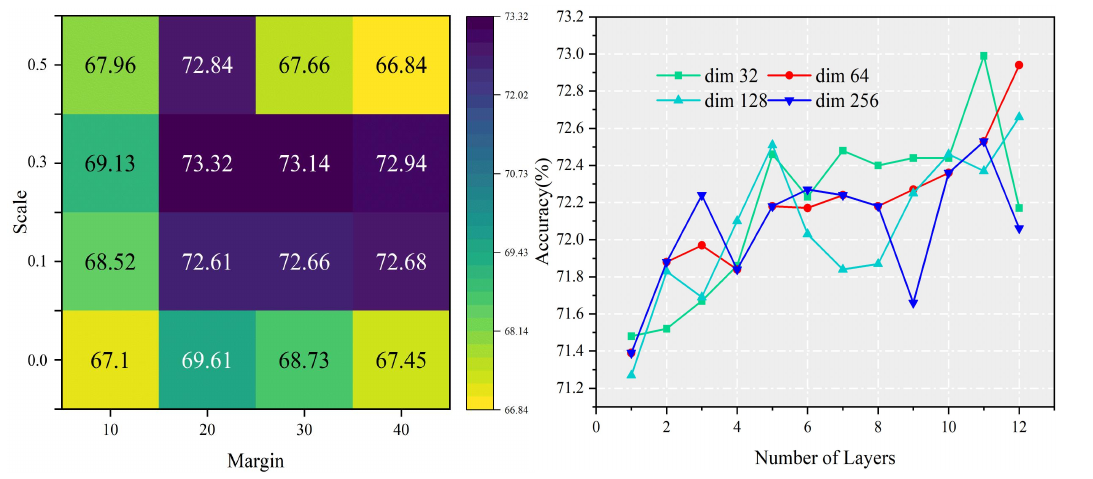}        
\caption{The results of some hyperparameters. The left shows the effect of different combinations of margin and scale on the ImageNetA dataset. The right demonstrates the impact of varying the number of inserted adapter layers and the dimensionality of the adapters on performance. }
\label{hyper visual}
\centering
\end{figure}

\textbf{Analysis of margin and scale:} We conduct experiments on the ImageNetA dataset on the hyper-parameter of the scale and margin in our cosine loss as shown in Fig.~\ref{hyper visual} left. We discover that appropriately increasing scale can enhance the performance of the model when the margin is fixed. For example, when the margin is fixed at $20$, the performance is observed to continuously improve as the scale parameter is varied from $0.0$ to $0.5$.
when the scale is set to $0.5$, the average accuracy is $3.23\%$ higher than when the value is $0.0$. We also conduct experiments to analyze the influence of the margin. As shown in Fig.~\ref{hyper visual} left. It can be observed that with the scale parameter held fixed, appropriately increasing the margin can also enhance the performance of the model.
we also observe that different datasets have different appropriate margin values. More experimental results are presented in the \textit{Supp}. 

\textbf{Adapter dimension and layers insertion:} As shown in Fig.~\ref{hyper visual} right, we further conduct experiments on the specific position to insert the adapter module on the ImageNetA dataset.  We observe that the performance is progressively improved with the number of layers increased when the middle hidden layer dimension is fixed. Thus, we insert adapter modules in all $12$ layers in our comparative experiments of various methods. We also conduct ablation experiments on the middle dimension in the adapter also in Fig.~\ref{hyper visual} right. We observe that increasing the dimension has a positive effect on the performance of the model when the position to insert adapter module is fixed.  Interestingly, setting the middle dimension to 32 did not result in a significant decrease in performance. On the other hand, setting it to 256 led to an improvement in performance but also quadrupled the number of parameters.  To strike a balance between performance and the number of fine-tuning parameters, we set the middle dimension to $64$.

\begin{table*}[hpt]
\centering
\caption{The CIL results of different semantic shift methods on ImageNetA.}
\scalebox{0.98}{
\begin{tabular}{cccccccccccccc}
\toprule
                  Type             &    Method      & Data Nums & Ses.1    & Ses.2    & Ses.3    & Ses.4    & Ses.5   & Ses.6   & Ses.7   & Ses.8   & Ses.9   & Ses.10   & Avg$\uparrow$    \\ \midrule
                               & CA\cite{zhang2023slca}       & All       & 85.14 & 82.50  & 77.52 & 73.35 & 71.58 & 68.72 & 64.50  & 64.26 & 61.65 & 60.96 & 71.01 \\
                               \hline
\multirow{3}{*}{Instance-Based}      & $l_1$ Distance \cite{tan2024semantically}    & All       & 85.14 & 82.50  & 77.31 & 77.22 & 75.51 & 72.10  & 65.74 & 66.11 & 64.52 & 62.41 & 72.85 \\
                               & $l_2$ Distance \cite{tan2024semantically}    & All       & 85.14 & 82.50  & 77.31 & 77.22 & 75.51 & 72.10  & 65.74 & 66.19 & 64.52 & 62.41 & 72.86 \\
                               & RoP-ILAP\cite{ma2023remind}  & k=100       & 85.14 & 82.50  & 77.31 & 70.92 & 66.35 & 60.51 & 53.43 & 49.84 & 46.95 & 44.63 & 63.75 \\
                               & RoP-ILAP\cite{ma2023remind}  & k=200       & 85.14 & 83.06 & 77.10  & 72.06 & 68.61 & 65.64 & 58.97 & 56.57 & 53.12 & 50.63 & 67.09  \\
                               & RoP-ILAP\cite{ma2023remind}  & k=300       & 85.14 & 83.06 & 77.73 & 73.93 & 71.22 & 67.79 & 61.20  & 60.98 & 57.28 & 54.84 & 69.31 \\
                               & RoP-ILAP\cite{ma2023remind}  & All       & 85.14 & 82.78 & 77.10  & 77.22 & 75.51 & 71.90  & 66.01 & 66.67 & 64.01 & 62.54 & 72.88 \\
                               \hline
\multirow{2}{*}{Prototype-Based} & PRAKA\cite{shi2023prototype} &     All      & 85.14 & 80.56 & 73.74 & 66.48 & 65.16 & 60.82 & 57.36 & 54.57 & 52.26 & 50.56 & 64.66 \\
                               & SSIAT(Ours) &  All         & 85.14 & 82.78 & 77.31 & 77.51 & 75.03 & 71.79 & 66.10  & 66.59 & 64.23 & 63.00 & 72.94 \\
\bottomrule
\end{tabular}}

\label{different ssca}
\end{table*}




\begin{table}[hpt]
\centering
\caption{Results for different PET methods on ImageNetR/A. }
\scalebox{0.88}{
\begin{tabular}{cccccc}
\toprule
 \multirow{2}*{Method} &\multicolumn{2}{c}{ImageNetR} &\multicolumn{2}{c}{ImageNetA}  \\
 & $\mathcal{A}_{Last}\uparrow$& $\mathcal{A}_{Avg}\uparrow$&$\mathcal{A}_{Last}\uparrow$ &$\mathcal{A}_{Avg}\uparrow$ \\
\hline
SSF   & $71.84_{\pm 0.33}$ & $79.98_{\pm 0.79}$ & $52.11_{\pm 0.64}$ & $62.34_{\pm 1.33}$ & 
              \\
+SSCA & $75.01_{\pm 0.31}$ & $82.09_{\pm 0.41}$  & $58.94_{\pm 1.09}$ & $67.94_{\pm 1.06}$ & \\
\hline
VPT-deep  & $38.49_{\pm 0.13}$  & $50.34_{\pm 1.93}$  & $37.39_{\pm 22.03}$  & $46.55_{\pm 16.69}$ &          \\
+SSCA & $56.11_{\pm 3.25}$  & $61.11_{\pm 1.71}$ & $47.83_{\pm 18.75}$ & $55.67_{\pm 14.92}$ &   \\
\hline
VPT-shallow   & $58.79_{\pm 1.07}$ & $69.23_{\pm 4.06}$  & $48.34_{\pm 0.99}$  & $56.96_{\pm 3.45}$ &          \\
+SSCA & $68.25_{\pm 2.50}$  & $72.40_{\pm 2.23}$ & $54.49_{\pm 0.76}$ & $62.26_{\pm 2.54}$ &   \\
\bottomrule
\end{tabular}}

\label{A and R}
\end{table}



\textbf{Different pre-trained models.} We experiment with pre-trained models (PTMs) with different generalization abilities on four datasets shown in Tab.~\ref{Different ability}. We conduct experiments on three different versions of ViT PTMs and it can be observed that our method generalizes well to various PTMs. Furthermore, The large-based ViT model can get better performance on ImageNetR and ImageNetA datasets. However, on other datasets, such as CIFAR100, the large-based ViT models may not necessarily achieve better performance.

\textbf{Different tuning methods with SSCA.} We observe that the adapter module exhibits the best performance with continuously fine-tuning. Therefore, we combined the semantic shift correction with continuous adapter fine-tuning, which led to even better results. To further substantiate the superior performance of the adapter, we also conduct experiments by integrating the semantic shift correction with other PET methods for comparison.
We incorporate classifier alignment with semantic shift estimation into SSF and prompt tuning shown in Tab.~\ref{A and R}. It can be seen that both the performance of prompt-based and SSF tuning approaches show significant improvement. However, our proposed method still outperforms them by a large margin. The results further verify the effectiveness of our proposed method.

\begin{figure}[t]
\centering
\includegraphics[scale=0.41]{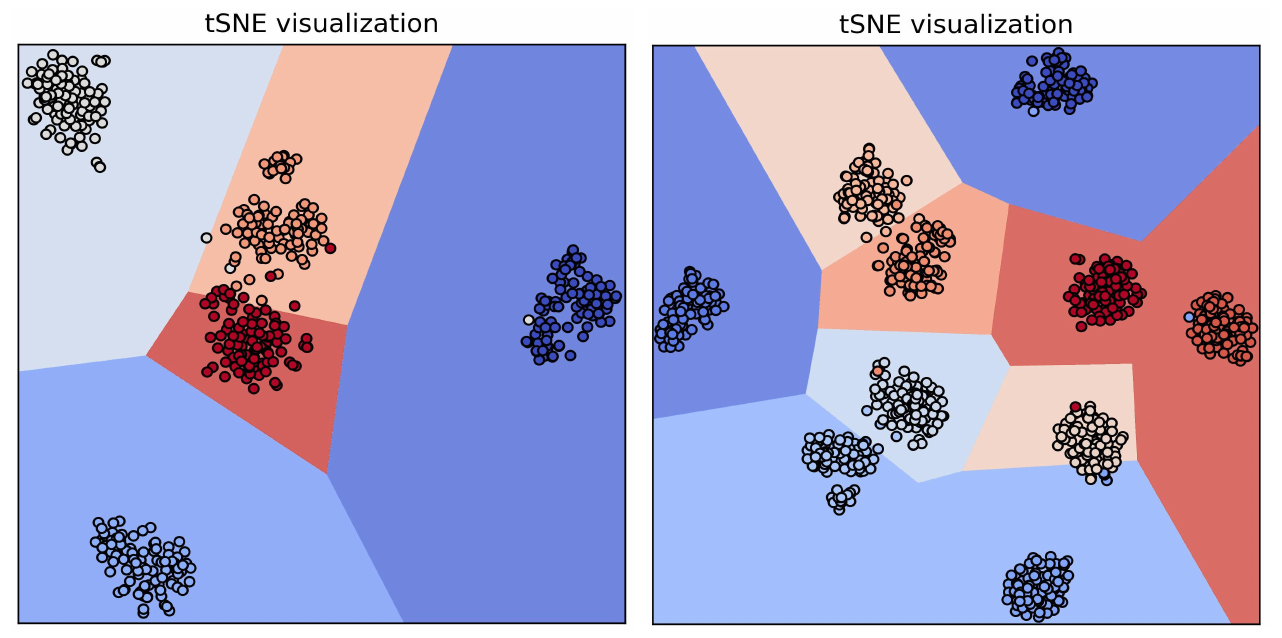}        
\caption{The results of feature and decision boundary visualization on CIFAR100 dataset. The left shows five classes in the base session and the right shows ten classes after an incremental session. }
\label{tsne}
\centering
\end{figure}

\textbf{Visualization of decision boundaries and confusion matrixes.} To visually observe the changes in class decision boundaries during the incremental process, we conduct feature visualization on the CIFAR100 dataset. The visualization results are shown in Fig.~\ref{tsne}. It can be seen that the discriminability of the base session features is excellent. After undergoing incremental training to accommodate the newly added classes, the discriminability of the new classes is also preserved, without affecting the discriminability of the base session classes. We show the confusion matrix on ImageNetA, CUB, and ImageNetR datasets in Fig.~\ref{confusion_matrix}. We show the results across four sessions for each dataset. It can be seen that regardless of whether the current session involves learning new classes or previously learned classes, the model does not appear to exhibit significant confusion in its predictions.

\begin{table}[htp]
\centering
\caption{Results on different pre-trained models on ImageNetR/A. }
\scalebox{0.86}{
\begin{tabular}{cccccc}
\toprule
 \multirow{2}*{Pre-trained Model} &\multicolumn{2}{c}{ImageNetR} &\multicolumn{2}{c}{ImageNetA}  \\
 & $\mathcal{A}_{Last}\uparrow$& $\mathcal{A}_{Avg}\uparrow$&$\mathcal{A}_{Last}\uparrow$ &$\mathcal{A}_{Avg}\uparrow$ \\
\hline
 ViT-base 1K   &$80.02_{\pm 0.30}$  &$84.45_{\pm 0.42}$&$61.40_{\pm 0.78}$ &  $71.05_{\pm 1.20}$  \\
 ViT-base 21k &$81.96_{\pm 2.60}$ &  $85.64_{\pm 1.24}$ &$60.78_{\pm 1.52}$ &  $70.00_{\pm 1.06}$   \\ 
 ViT-large 21k &$80.02_{\pm 0.30}$ & $84.12_{\pm 0.21}$  & $65.55_{\pm 3.76}$ & $74.12_{\pm 3.16}$ \\
\hline
\multirow{2}*{Pre-trained Model} &\multicolumn{2}{c}{CIFAR100} &\multicolumn{2}{c}{CUB200}  \\
& $\mathcal{A}_{Last}\uparrow$& $\mathcal{A}_{Avg}\uparrow$&$\mathcal{A}_{Last}\uparrow$ &$\mathcal{A}_{Avg}\uparrow$ \\
\hline
ViT-base 1K   &$90.68_{\pm 0.23}$  &$93.72_{\pm 1.13}$&$88.32_{\pm 0.48}$ &  $92.94_{\pm 0.55}$  \\
ViT-base 21k &$91.02_{\pm 0.87}$ &  $94.24_{\pm 0.88}$ &$89.82_{\pm 0.59}$ &  $93.76_{\pm 0.58}$   \\ 
 ViT-large 21k &$90.63_{\pm 0.69}$ & $92.77_{\pm 1.96}$  & $90.23_{\pm 0.05}$ & $93.62_{\pm 0.14}$ \\
 \bottomrule
\end{tabular}}

\label{Different ability}
\end{table}

\textbf{Different methods of SSCA.} There exist multiple methods available to calculate the similarity between features in SSCA. To compare the effectiveness of different SSCA approaches, we conduct experiments on ImageNetA dataset and the results are shown in Tab.~\ref{different ssca}. It can be seen that when incorporating the currently available data to correct semantic drift, various methods for computing similarity yield similar results. RoP-ILAP \cite{ma2023remind} utilizes the $k$-means algorithm to identify samples closest to the existing class prototypes and then calculates the corresponding semantic shift. We conduct experiments using the same method with varying values of $k$. The results indicate that as the value of $k$ increased, the average performance improved. This shows that incorporating all samples leads to better alleviation of semantic drift. Furthermore, it can be seen that our proposed prototype-based method yields better results compared to using the entire sample set. PRAKA employs an interpolation-based sampling technique to correct prototypes using features from the old class. However, the results indicate that its performance is moderate on imbalanced datasets such as ImageNetA.

\section{Conclusion}
\label{conclusion}
Class-incremental learning on a pre-trained model has received significant attention and gained excellent performance in recent years. In this paper, we first revisit different PET methods in the context of CIL. Then, we propose that incrementally tuning the shared adapter and local classifier without constraints exhibits less forgetting and gains plasticity for learning new classes. Moreover, to train a unified classifier, we calculate the semantic shift of old prototypes and retrain the classifier using updated prototypes without access to past image samples in each session. The proposed method eliminates the need for constructing an adapter pool and avoids retaining any image samples. Experimental results on several CIL benchmarks and more general settings demonstrate the effectiveness of our method which achieves the SOTA performance. 

\textbf{Limitation}. 
We demonstrate that the PET method offers significant advantages in CIL with pre-trained models, and adapter-based methods generally outperform other PET-based methods. However, further validation is needed for PET methods in real-world open-domain scenarios. Additionally, addressing semantic shift and retraining the classifier is required at each incremental session, leading to substantial time overhead as the number of sessions increases. Although our prototype-based semantic shift method reduces time overhead compared to other approaches, its application remains limited in high real-time requirements. Beyond ViT-based pre-trained models, research is increasingly focusing on vision-language multimodal pre-trained models for incremental learning tasks. Enhancing the performance of these multimodal models in more complex incremental learning settings will likely be a key area for future work.


%


\bibliographystyle{IEEEtran}
\bibliography{main}

\ifCLASSOPTIONcaptionsoff
  \newpage
\fi

\end{document}